\documentclass[10pt,twocolumn,letterpaper]{article}

\usepackage{iccv}
\usepackage{times}
\usepackage{epsfig}
\usepackage{graphicx}
\usepackage{amsmath}
\usepackage{amssymb}

\usepackage{microtype}
\usepackage{graphicx}
\usepackage{booktabs} % for professional tables

\usepackage[numbers]{natbib}
\usepackage[ruled,boxed,algo2e]{algorithm2e}
\usepackage[title]{appendix}
\usepackage{multicol}
\usepackage{bm}
\usepackage{amsmath}
\usepackage{amssymb}
\usepackage{multirow}
\usepackage{graphicx}
\usepackage{xcolor}
\usepackage{subcaption}
\usepackage{wrapfig}

\newcommand{\ourmethod}{\emph{Rails}}

\newcommand{\reffig}[1]{Figure~\ref{fig:#1}}

\newcommand{\refsec}[1]{Section~\ref{sec:#1}}

\newcommand{\reftbl}[1]{Table~\ref{Tbl:#1}}
\newcommand{\refalg}[1]{Algorithm~\ref{alg:#1}}

\newcommand{\shortrefsec}[1]{\S\ref{sec:#1}}
\newcommand{\refeq}[1]{Equation~\eqref{eq:#1}}

\newcommand{\lblfig}[1]{\label{fig:#1}}
\newcommand{\lblsec}[1]{\label{sec:#1}}
\newcommand{\lbleq}[1]{\label{eq:#1}}

\newcommand{\lbltbl}[1]{\label{Tbl:#1}}
\newcommand{\lblalg}[1]{\label{alg:#1}}

\usepackage[pagebackref=true,breaklinks=true,letterpaper=true,colorlinks,bookmarks=false]{hyperref}

\iccvfinalcopy % *** Uncomment this line for the final submission

\ificcvfinal\fi

\begin{document}

%%%%%%%%% TITLE
\title{Learning to drive from a world on rails}

\author{Dian Chen\\
UT Austin
\and
Vladlen Koltun\\
Intel Labs
\and
Philipp Kr\"ahenb\"uhl\\
UT Austin
}
\maketitle
\ificcvfinal\thispagestyle{empty}\fi

%%%%%%%%% ABSTRACT
\begin{abstract}
We learn an interactive vision-based driving policy from pre-recorded driving logs via a model-based approach. A forward model of the world supervises a driving policy that predicts the outcome of any potential driving trajectory. To support learning from pre-recorded logs, we assume that the world is on rails, meaning neither the agent nor its actions influence the environment. This assumption greatly simplifies the learning problem, factorizing the dynamics into a nonreactive world model and a low-dimensional and compact forward model of the ego-vehicle. Our approach computes action-values for each training trajectory using a tabular dynamic-programming evaluation of the Bellman equations; these action-values in turn supervise the final vision-based driving policy. Despite the world-on-rails assumption, the final driving policy acts well in a dynamic and reactive world. At the time of writing, our method ranks first on the CARLA leaderboard, attaining a $25\%$ higher driving score while using $40\times$ less data.
Our method is also an order of magnitude more sample-efficient than state-of-the-art model-free reinforcement learning techniques on navigational tasks in the ProcGen benchmark.
 
\end{abstract}

\section{Introduction}

Vision-based autonomous driving is hard.
An agent needs to perceive, understand, and interact with its environment from incomplete and partial experiences.
Most successful driving approaches~\cite{chen2020learning,mueller2018driving,sadat2020perceive,sauer2018conditional} reduce autonomous navigation to imitating an expert, usually a human actor.
Expert actions serve as a source of strong supervision, sensory inputs of the expert trajectories explore the world, and policy learning reduces to supervised learning backed by powerful deep networks.
However, expert trajectories are often heavily biased, and safety-critical observations are rare.
After all, human operators drive hundreds of thousands of miles before observing a traffic incident~\cite{tefft2017rates}. % Estimated 250k miles between accident CITE: Driving to safety: How many miles of driving would it take to demonstrate autonomous vehicle reliability?
This sparsity of safety-critical training data makes it difficult for a behavior-cloning agent
to learn and recover from mistakes.
Model-free reinforcement learning~\cite{liang2018cirl, toromanoff2020end} offers a solution, allowing an agent to actively explore its environment and learn from it.
However, this exploration is even less data-efficient than behavior cloning, as it needs to experience mistakes to avoid them.
For reinforcement learning, the required sample complexity for safe driving is prohibitively large, even in simulation~\cite{toromanoff2020end}. 

\begin{figure}[t]
\includegraphics[width=\linewidth]{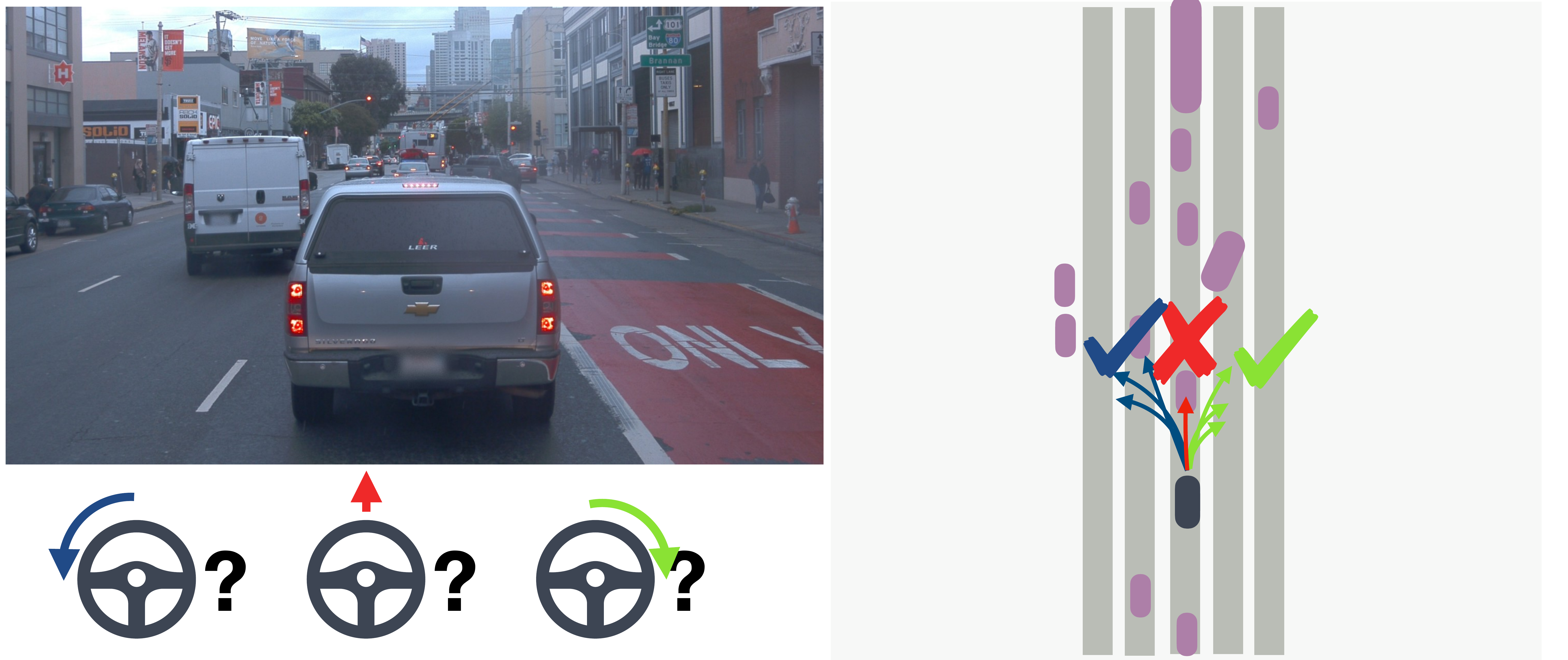}
\caption{We learn a reactive visuomotor driving policy that gets to explore the effects of its own actions at training time. The policy simulates the effects of its own actions using a forward model in pre-recorded driving logs. It then learns to choose safe actions without explicitly experiencing unsafe driving behavior. Picture selected from the Waymo open dataset~\cite{sun2020scalability}. 
}
\lblfig{teaser}
\end{figure}

\begin{figure*}[t]
    \centering
    \begin{subfigure}[b]{0.2\linewidth}
     \centering
      \includegraphics[width=\textwidth,page=1]{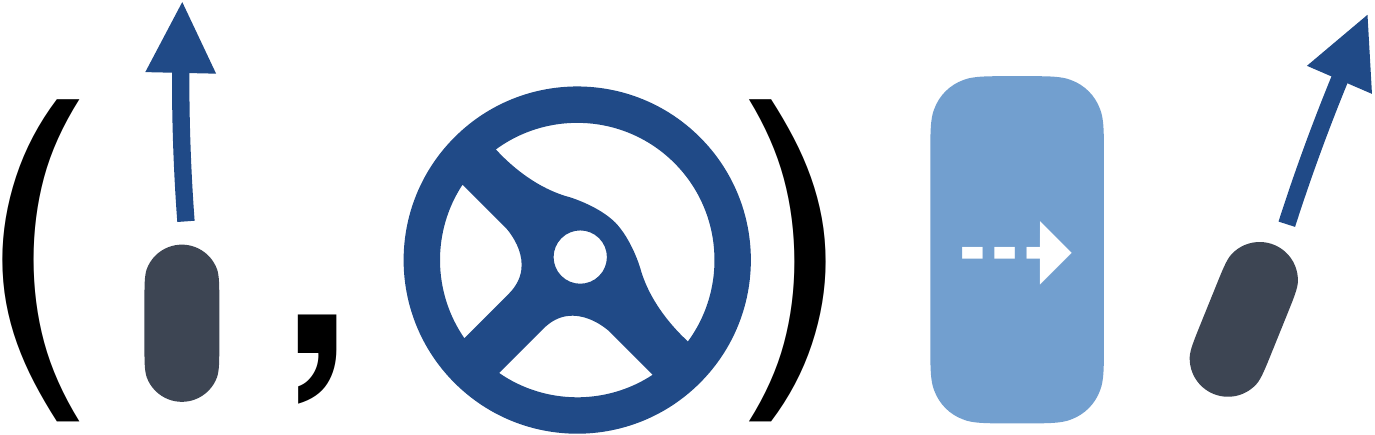}
     \caption{Forward model}
    \end{subfigure}\hfill
    \begin{subfigure}[b]{0.34\linewidth}
     \centering
      \includegraphics[width=\textwidth,page=2]{figures/overview2_crop.pdf}
     \caption{Bellman update}
    \end{subfigure}\hfill
    \begin{subfigure}[b]{0.4\linewidth}
     \centering
      \includegraphics[width=\textwidth,page=3]{figures/overview2_crop.pdf}
     \caption{Distillation}
    \end{subfigure}%
      \caption{Overview of our approach. Given a dataset of offline driving trajectories of sensor readings, driving states, and actions, we first learn a forward model of the ego-vehicle (a). Using the offline driving trajectories, we then compute action-values under a predefined reward and learned forward model using dynamic programming and backward induction on the Bellman equation (b). Finally, the action-values supervise a reactive visuomotor driving policy through policy distillation (c). 
      For a single image, we supervise the policy for all vehicle speeds and actions for a richer supervisory signal. % This is carla challenge only yetVsd
    }
\lblfig{overview}
\end{figure*}

In this paper, we present a method to learn a navigation policy that recovers from mistakes without ever making them, as illustrated in \reffig{teaser}.
We first learn a world model on static pre-recorded trajectories.
This world model is able to simulate the agent's actions without ever executing them.
Next, we estimate action-value functions for all pre-recorded trajectories.
Finally, we train a reactive visuomotor policy that gets to observe the impact of all its actions as predicted by the action-value function.
The policy learns to avoid costly mistakes, or recover from them.
We use driving logs, recorded lane maps and locations of traffic participants, to train the world model and compute the action-value function.
However, our visuomotor policy drives using raw sensor inputs, namely RGB images and speed readings alone.
\reffig{overview} provides an overview.

The core challenge in our approach is to build a sufficiently expressive and accurate world model that allows the agent to explore its environment and the impact of its actions.
For autonomous driving, this involves modeling the autonomous vehicle and all other scene elements, such as other vehicles, pedestrians, traffic lights, etc.
In its raw form, the state space in which the agent operates is too high-dimensional to effectively explore.
We thus make a simplifying assumption: The agent's actions only affect its own state, and cannot directly influence the environment around it.
In other words: the world is ``on rails''.
This naturally factorizes the world model into an agent-specific component that reacts to the agent's commands, and a passively moving world.
For the agent, we learn an action-conditional forward model.
For the environment, we simply replay pre-recorded trajectories from the training data.

The factorization of the world model lends itself to a simple evaluation of the Bellman equations through dynamic programming and backward induction.
For each driving trajectory, we compute a tabular approximation of the value function over all potential agent states.
We use this value function and the agent's forward model to compute action-value functions, which then supervise the visuomotor policy.
The action values are computed over all agent states, and thus serve as denser supervision signals for the same number of environment interactions. They provide the visuomotor policy with action targets for any camera viewpoint, vehicle speed, or high level command augmentations.

We evaluate our method in the CARLA simulator~\cite{dosovitskiy17}.
On the CARLA leaderboard\footnote{\url{https://leaderboard.carla.org/leaderboard/}}, we achieve a $25\%$ higher driving score than the prior top-ranking entry while using $40\times$ less training data. Notably, our method uses camera-only sensors, while some prior work relies on LiDAR.
We also outperform all prior methods on the NoCrash benchmark~\cite{codevilla2019exploring}.
Finally, we show that our method generalizes to other environments using the ProcGen platform~\cite{cobbe2019leveraging}.
Our method successfully learns navigational policies in the \textit{Maze} and \textit{Heist} environments with an order of magnitude fewer observations than baseline algorithms. Code and data are available\footnote{ \url{https://dotchen.github.io/world_on_rails}}.

\section{Related Work}

\textbf{Imitation learning} 
is one of the earliest and most successful approaches to vision-based driving and navigation.
\citet{pomerleau1989alvinn} pioneered this direction with ALVINN.
Recent work extends imitation learning to challenging urban driving and navigation in complicated environments \cite{muller2006off,pan2018agile,bansalchauffeurnet,codevilla2018,codevilla2019exploring,sauer2018conditional,liang2018cirl}.
Imitation learning algorithms train on trajectories collected by human experts~\cite{codevilla2018,dosovitskiy17,pomerleau1989alvinn}, or privileged experts constructed with rich sensory data~\cite{chen2020learning,pan2018agile}. These approaches are limited to the expert's observations and actions. In contrast, our work learns to drive from passive driving logs and integrates mental exploration into the learning process so as to imagine and learn from scenarios that were not experienced when the logs were collected.

\textbf{Model-based reinforcement learning} builds a forward model to help train the policy.
\citet{sutton1991planning, gu2016continuous, kalweit2017uncertainty, kurutach2018model} use a forward world model to generate imagined trajectories to improve the sample complexity.
World models~\cite{oh2017value, ha2018worldmodels, hafner2019dream, schrittwieser2020mastering} use the forward model to provide additional context to assist the learning agents' decision making.
\citet{feinberg2018model, buckman2018sample} roll out the forward model for short horizons to improve the fidelity of their Q or value function approximation.
In our work, we factorize the forward world model into the controllable ego-agent and a passively moving environment.
This factorization significantly simplifies policy learning and allows for a tabular evaluation of the Q and value functions. Our idea of factorizing the agent and the environment is similar to the idea of \textit{exogenous events} in policy learning \cite{boutilier1999decision}.
Recently, \citet{dietterich2018discovering,chitnis2020learning} considered finding a minimal factorized MDP.
In contrast, we explicitly factorize the environment and focus on leveraging the factorization for planning and supervision of a visuomotor policy. 

\textbf{Policy distillation} remaps the outputs of a privileged agent to a visuomotor agent~\cite{chen2020learning,levine2016end,pan2018agile,lee2020learning}.
\citet{levine2016end} use optimal control methods to learn local controllers for robotic manipulation tasks, and use them to supervise a visuomotor policy.
\citet{pan2018agile} train a visuomotor driving policy by imitating an MPC controller that has access to expensive sensors.
\citet{lee2020learning} first learn a privileged policy using model-free RL, then distill a visuomotor agent.
\citet{chen2020learning} distill a visuomotor agent from a policy learned by imitation on privileged simulator states.
Our approach uses a similar privileged simulator state to infer an action-value function to supervise the final visuomotor policy.
While prior work uses one policy to supervise another, in our work a tabular action-value function supervised the policy.
A reactive driving policy only exists after distillation.

\textbf{Cost volume} based planners~\cite{zeng2019end,sadat2020perceive,casas2021mp3} score and rank select future ego-vehicle trajectories.
In tabular form, they closely resemble our action-value estimate.
However, our action-value estimate has two advantages.
First, it supervises a policy at training time in an offline process, while cost volumes need to be predicted for inference~\cite{zeng2019end,sadat2020perceive,casas2021mp3}.
Second, we make use of ground-truth states, while cost volumes use imitation~\cite{zeng2019end} or affordances~\cite{sadat2020perceive,casas2021mp3} from partial observations.

\section{Method}

We aim to learn a reactive visuomotor policy $\pi(I)$ that produces an action $a \in \mathcal{A}$ for a sensory input $I$.
At training time, we are given a set of trajectories $\tau \in D$.
Each trajectory $\tau=\{(\hat I_1, \hat L_1, \hat a_1), (\hat I_2, \hat L_2, \hat a_2), \ldots\}$ contains a stream of sensor readings $\hat I_t$, corresponding driving logs $\hat L_t$, and executed actions $\hat a_t$.
The hat symbols denotes data from driving logs, regular symbols denote free or random variables.
The driving logs record the state (position, velocity, and orientation) of the ego-vehicle and all other traffic participants, as well as the environment state (lane information, traffic light state, etc.).
We use the driving logs to compute a forward model $\mathcal{T}$ of the world and an action-value function $Q$ from a scalar reward.
The forward model $\mathcal{T}$ takes a driving state $L_t$ and an agent's action $a_t$ to predict the next state $L_{t+1}$.
We use a hybrid semi-parametric model to estimate $\mathcal{T}$, as described in \refsec{forward_model}.
Specifically, we factorize the forward model into a ego-vehicle component $\mathcal{T}^{ego}$ and a world component $\mathcal{T}^{world}$.
We approximate the ego-vehicle forward model using a simple deep network, while the collected trajectories are used non-parametrically for the world forward model.
This factorization allows us to estimate an action-value function using a tabular approximation of the Bellman equation, as described in \refsec{value_iteration}.
Finally, we use the estimated action-values $Q$ to distill a visuomotor policy $\pi$.
This policy $\pi$ maximizes the expected return under our forward model and tabular action-value approximation.
At training time, our algorithm uses privileged information, i.e. driving logs, to supervise policy learning, but the final policy $\pi(I_t)$ drives from sensor inputs alone.
\refalg{overview} and \reffig{overview} summarize the entire training process.

\setlength{\algomargin}{0.3em} 
\begin{algorithm2e}[t]
 \SetAlgoLined
 \SetKwProg{Fn}{Function}{:}{end}
 \newcommand{\fname}{\mathbf}
 \KwData{Training trajectories $D$}
 \KwResult{Policy $\pi(I) \in \mathcal{A}$}
 \tcp{Forward-model fitting \shortrefsec{forward_model}}
 \Fn{$\fname{FitForward}(D) \to \mathcal{T}^{ego}$}{
%   Collect training data $\mathcal{D} = \{(\hat L_t, \hat a_t, \hat L^{ego}_{t+1}) | \hat L_t \in \tau, \tau \in D\}$\;
   Minimize \refeq{forward}\;
   \textbf{return} ego-vehicle forward model $\mathcal{T}^{ego}$\;
 }
 \tcp{Action-value estimate \shortrefsec{value_iteration}}
 \Fn{$\fname{EstimateQ}(D, \mathcal{T}^{ego}) \to Q$}{
   \For{$\tau \in D$}{
     Initialize $V_{|\tau|+1}(\cdot) = 0$\;
     \For{$t = |\tau| \ldots 1$}{
       Compute $Q_t$ an $V_t$ \refeq{bellman}\;
       Store $Q_t$\;
     }
   }
   \textbf{return} stored $Q$-values\;
 }
 \tcp{Policy distillation \shortrefsec{policy}}
 \Fn{$\fname{DistillPolicy}(D, Q) \to \pi$}{
   Minimize \refeq{distill}\;
   \textbf{return} visuomotor policy $\pi$\;
 }
 \makebox[10em]{Learn forward model\hfill}$\mathcal{T}^{ego}\!=\!\fname{FitForward}(D)$\;
 \makebox[10em]{Estimate action-values\hfill}$Q\!=\!\fname{EstimateQ}(D, \mathcal{T}^{ego})$\;
 \makebox[10em]{Learn visuomotor policy\hfill}$\pi\!=\!\fname{DistillPolicy}(D, Q)$\;
 \caption{Learning in a world-on-rails}
 \lblalg{overview}
\end{algorithm2e}

\subsection{A factorized forward model}
\lblsec{forward_model}
In its raw form the forward model $\mathcal{T}$ is too complex to efficiently predict and simulate.
After all, entire driving simulators are designed to forecast just one of the many possible future driving states.
We thus factorize the driving state $L_t$ and forward model $\mathcal{T}$ into two parts: A part considering just the vehicle being controlled ${L^{ego}_{t+1} = \mathcal{T}^{ego}(L^{ego}_t, L^{world}_t, a_t)}$ and a part modeling the rest of the world ${L^{world}_{t+1} = \mathcal{T}^{world}(L^{ego}_t, L^{world}_t, a_t)}$.
Here we consider only deterministic transitions.
We furthermore assume that \emph{the world is on rails} and cannot react to the agents' commands $a$ or the state of the ego-vehicle $L^{ego}$.
Specifically, the transition of the world state only depends on the prior world state itself: ${L^{world}_{t+1} = \mathcal{T}^{world}(L^{world}_t)}$.
Thus the initial state of the world $L^{world}_{0}$ determines the entire trajectory of the world: $\{L^{world}_{1}, L^{world}_{2}, \ldots\}$.
This allows us to model the world transition using the collected trajectories $\tau$ directly.
We thus only need to model the ego-vehicle's forward-model $\mathcal{T}^{ego}$ for any ego-vehicle state $L^{ego}_t$ and action $a_t$.
We train $\mathcal{T}^{ego}$ on the collected trajectories using L1 regression
\begin{equation}
  E_{\hat L^{ego}_{t:t+T}\!,\hat a_t}\!\!\left[\sum_{\Delta=1}^T\left|{\mathcal{T}^{ego}}^\Delta(\hat L^{ego}_{t}\!, \hat a_{t+\Delta-1})\!-\!\hat L^{ego}_{t+\Delta}\right|\right]\!,\lbleq{forward}
\end{equation}
where we roll out the forward model for $T=10$ steps to obtain a more robust regression target.
We use a simple parametric bicycle model that easily generalizes beyond the training states $\hat L^{ego}$, as described in \refsec{implementation}.

The world-on-rails assumption clearly does not hold, neither in a simulator nor in the real world.
Other agents in the world \emph{will} react to the ego-vehicle and its actions.
However, this does not imply that a world-on-rails cannot provide strong and useful supervision to the agent.
Our experiments show that an agent trained in a world-on-rails significantly outperforms agents trained with a full forward model of the world.
The world-on-rails assumption significantly simplifies the estimation of an action-value function in \refsec{value_iteration} and subsequent policy learning in \refsec{policy}.

\subsection{A factorized Bellman equation}
\lblsec{value_iteration}

Our goal is to estimate an action-value function $Q(\hat L_t, a)$ for each state $\hat L_t$ of the training trajectory and action $a$.
We use the Bellman equation and a tabular discretization of the value function here.
Recall the $\gamma$-discounted Bellman equation: $V(L_t) = \max_a Q(L_t, a)$ and $Q(L_t, a) = \gamma V(\mathcal{T}(L_t,a)) + r(L_t, a)$ for any state $L_t$, action $a$, and reward $r$.
Ordinarily, one would need to resort to Bellman iterations to estimate $V$ and $Q$.
However, our factorized forward-model simplifies this:
\begin{align*}
  V(L^{ego}_t\!, \hat L^{world}_t) =& \max_a Q(L^{ego}_t\!, \hat L^{world}_t\!, a)\\
  Q(L^{ego}_t\!, \hat L^{world}_t\!, a_t) =& r(L^{ego}_t\!, \hat L^{world}_t\!, a_t)+\\
  &\gamma V(\mathcal{T}^{ego}(L^{ego}_t\!, \hat L^{world}_t\!, a),\hat L^{world}_{t+1}).\notag
\end{align*}
The action-value function is needed for all ego-vehicle state $L^{ego}$, but only recorded world states $\hat L_t^{world}$.
It is sufficient to evaluate the action-value function on just recorded world states for all ego-vehicle states: $\hat V_t(L^{ego}_t) = V(L^{ego}_t\!, \hat L^{world}_t)$, $\hat Q_t(L^{ego}_t, a_t) = V(L^{ego}_t\!, \hat L^{world}_t, a_t)$.
Furthermore, the world states are strictly ordered in time, hence
the Bellman equations simplifies to
\begin{align}
  \hat{V}_t(L^{ego}_t) =& \max_a \hat{Q}_t(L^{ego}_t\!, a)\lbleq{bellman}\\
  \hat{Q}_t(L^{ego}_t\!, a_t) =& r(L^{ego}_t\!, \hat L^{world}_t\!, a_t)+\notag\\
  & \gamma \hat{V}_{t+1}(\mathcal{T}^{ego}(L^{ego}_t\!, a)).\notag
\end{align}
Here the value and action-value functions only consider recorded world states, but all possible ego-vehicle states.
The model is thus able to ``imagine'' driving behaviors and their reward without ever executing them.
In order to collect rewards from these ``imagined'' states, we require an explicit reward function $r$, and not just a scalar reward signal provided by the environment.
For a detailed discussion of the reward see \refsec{reward}.

We solve \refeq{bellman} using backward induction and dynamic programming.
The state of the ego-vehicles $L^{ego}$ is compact (position, orientation, and velocity).
This allows us to compute a tabular approximation of the value function $V_t(L^{ego}_t)$, evaluated in batch operations efficiently.
Specifically, we discretize $V_t(L^{ego}_t)$ into bins corresponding to the position, orientation, and velocity of the ego-vehicle.
When evaluating, we use linear interpolation if the requested value falls between bins.
Furthermore, the action space is also small, allowing for a discretization of the $\max$ operator in the value update.
During backward induction, we implicitly represent the action-value function $Q_t$ using $V_{t+1}$ and the forward model $\mathcal{T}^{ego}$.
We only discretize $Q_t(\hat L^{ego}_t, \cdot)$ to supervise the visuomotor policy at timestep $t$.
\refalg{overview} summarizes our backward induction. More details are provided in the supplementary material for reference.

\subsection{Policy Distillation}
\lblsec{policy}
We use the action-value functions for the ego-vehicle state $Q_t(\hat L^{ego}_t, \cdot)$ to supervise a visuomotor policy $\pi(\hat I_t)$.
The action-value $Q_t(\hat L^{ego}_t, \cdot)$ represents the expected return of an optimal policy each vehicle state.
We directly optimize this expected return in our policy:
\begin{equation}
  E_{\hat{L}^{world}_t, L^{ego}_t, \hat I_t}\left[ \sum_a \pi(a|\hat I_t) \hat{Q}_t(L^{ego}_t, a) + \alpha H\!\left(\pi(\cdot|\hat I_t)\right) \right]. \lbleq{distill}
\end{equation}
Since the action-value functions are computed densely, only the environment needs to be recorded, not the ego state. We can therefore supervise with augmented $\hat{I}_{t}$ representing arbitrary $L^{ego}_t$.
We additionally add an entropy regularizer $H$~\cite{haarnoja2018soft} to encourage a more diverse output policy, where $\alpha$ is the temperature hyperparameter.
In practice, we discretize both the action-values and the visuomotor policy below.

\section{Implementation}
\lblsec{implementation}

\paragraph{Forward model.}
\lblsec{forward}
We train the ego-vehicle forward model $\mathcal{T}^{ego}$ on a small subset of trajectories.
We collect the subset of trajectories to span the entire action space of the ego-vehicle: steering $s \in [-1,1]$ and throttle $t \in [0,1]$ are uniformly sampled, with brake $b \in \{0,1\}$ sampled from a Bernoulli distribution.
The forward model $\mathcal{T}^{ego}$ takes as inputs the current ego-vehicle state as 2D location $x_t,y_t$, orientation $\theta_t$, speed $v_t$, and predicts the next ego-vehicle state $x_{t+1}, y_{t+1},\theta_{t+1}, v_{t+1}$.
We use a parameterized bicycle model as the structural prior for $\mathcal{T}^{ego}$.
In particular, we only learn the vehicle wheelbases $f_b, r_b$, the mapping from user steering $s$ to wheel steering $\phi$, and the mapping from throttle and braking to acceleration $a$. The kinematics of the bicycle model are described in the appendix for reference.
We train $\mathcal{T}^{ego}$ in an auto-regressive manner using L1 loss and stochastic gradient descent.

\paragraph{Bellman equation evaluation.}
For each time-step $t$, we represent the value function $V_t$ as a 4D tensor discretized into $N_H\times N_W$ position bins, $N_v$ velocity bins, and $N_\theta$ orientation bins.
We use $N_H = N_W = 96$, $N_v=4$, and $N_\theta=5$.
Each bin has a physical size of $\frac{1}{3} \times \frac{1}{3} m^2$ and corresponds to a $2m/s$ velocity range and a $38^{\circ}$ orientation range.
The ego-vehicle state $\hat L^{ego}_t = (x_t, y_t, v, \theta)$ is always centered in this discretization.
The position of the ego-vehicle $(x_t, y_t)$ is at the center of the spatial discretization.
We only represent orientations in the range $[-95^{\circ},95^{\circ}]$ relative to the ego-vehicle.
When computing the action value function, any value $V_t$ that does not lie in the center of a bin is interpolated among its $2^4$ neighboring bins using linear interpolation. The linear interpolation is computed over all states at once and is factorized over ego state dimensions (location, speed and orientation), thus it is efficient.
Values that fall outside the discretization are $0$.
We discretize actions into $M_s \times M_t$ bins for steering and throttle respectively, and one additional bin for braking.
We do not steer or throttle while braking.
We use $M_s=9$ and $M_t=3$ for a total of $9\cdot3+1=28$ discrete actions. 

\paragraph{Policy network.}
The policy network uses a ResNet34~\cite{he2016deep} backbone to parse the RGB inputs.
We use global average pooling to flatten the ResNet features, before concatenating them with the ego-vehicle speed and feeding this to a fully-connected network.
The network produces a categorical distribution over the discretized action space.

In CARLA, the agent receives a high-level navigation command $c_t$ for each time-step.
% We include this navigation command in the agent stage $\hat L^{ego}$, add it to the discretization of the value function, and conditional the policy on this state.
We supervise the visuomotor agent simultaneously on all the high-level commands~\cite{chen2020learning}.
Additionally, we task the agent to predict semantic segmentation as an auxiliary loss.
This consistently improves the agent's driving performance, especially when generalizing to new environments.
We use image data augmentations following~\cite{codevilla2019exploring,chen2020learning}. More details on the augmentations are in the appendix for reference.

\paragraph{Reward design.}
\lblsec{reward}
The reward function $r(L_t^{ego}, L_t^{world}, a_t, c_t)$ considers ego-vehicle state, world state, action, and high-level command, and is computed from the driving log at each timestep.
We use the lane information of the world and high-level command to first compute the target lane of the ego-vehicle.
The agent receives a reward of $+1$ for staying in the target lane at the desired position, orientation and speed, and is smoothly penalized for deviating from the lane down to a value of 0.
If the agent is located at a ``zero-speed'' region (e.g.\ red light, or close to other traffic participants), it is rewarded for zero velocity regardless of orientation, and penalized otherwise except for red light zones. All ``zero speed'' rewards are scaled by $r_{\text{stop}}=0.01$, in order to avoid agents disregarding the target lane.
The agents receives a greedy reward of $r_{\text{brake}}=+5$ if it brakes in the zero-speed zone. To avoid agents chasing braking region, the braking reward cannot be accumulated.
All rewards are additive.
We found that with zero-speed zones and brake rewards, there is no need to explicitly penalize collisions. We compute the action-values over all high-level commands (``turn left'', ``turn right'', ``go straight'', ``follow lane'', ``change left'' or ``change right'') for each timestep, and use multi-branch supervision~\cite{chen2020learning} when distilling the visuomotor agent.

\section{Experiments}
\lblsec{dataset}
\paragraph{Dataset.}
We evaluate our approach on the open-source CARLA simulator~\cite{dosovitskiy17}.
We train our ego-vehicle forward model on a small subset of trajectories consisting of $2400$ collected frames.
It learns from random actions.

The bulk of our training set uses just passive sensor information $I$ and training logs $L$.
For the CARLA leaderboard, we collect $1M$ frames, corresponding to roughly 69 hours of driving.
For the NoCrash benchmark~\cite{codevilla2019exploring}, we collect $270K$ frames.
The dataset uses a privileged autopilot $\pi_b$.
However, we do not store the controls from the ego-vehicle autopilot, unlike imitation learning.
The RGB image is collected and stitched from three front-facing cameras all mounted at x=$1.5$m, z=$2.4$m in the ego-vehicle frame. 
Each camera has a $60^\circ$ FOV; the side cameras are angled at $55^\circ$. For the CARLA leaderboard, we additionally use a telephoto camera with $50^\circ$ FOV to capture distant traffic lights.
To augment the dataset, we additionally mount two side camera suites with the same setup, each mounted as if the vehicle is angled at $\pm 30^\circ$ following~\citet{bojarski2016end}.
For the CARLA leaderboard, we collect our dataset in the 8 public towns under a variety of weathers.
For the NoCrash benchmark, we collect our entire dataset in Town1 under four training weathers, as specified by the CARLA benchmark~\cite{dosovitskiy17,codevilla2018}. 

\paragraph{Experimental setup.}
We evaluate our approach on both the CARLA leaderboard and the NoCrash benchmark.
For both benchmarks, at each frame, the agent receives RGB camera reading $I$, speed reading $v$, and a high-level command $c$ to compute steering $s$, throttle $t$, and brake $b$.

For the CARLA leaderboard, agents are asked to navigate to specified goals through a variety of areas, including freeways, urban scenes, and residential districts, and in a variety of weather conditions.
The agents face challenging traffic situations along the route, including lane merging/changing, negotiations, traffic lights, and interactions with pedestrians and cyclists.
Agents are evaluated in held-out towns in terms of a Driving Score metric that is determined by route completion and traffic infractions.

In the NoCrash benchmark, agents are asked to safely navigate to specified goals in an urban setting with intersections, traffic lights, pedestrians, and other vehicles in the environment.
The NoCrash benchmark consists of three driving conditions, with traffic density ranging from empty to heavily packed with vehicles and pedestrians.
Each driving condition has the same set of 50 predefined routes: 25 in the training town (Town1) and 25 in an unseen town (Town2).
Agents are evaluated based on their success rates. A trial on a route is considered successful if the agent safely navigates from the starting position to the goal within a certain time limit.
The time limit corresponds to the amount of time required to drive the route at a cruising speed of $5$ km/h, excluding time spent stopping for traffic lights or other traffic participants.
In addition, a trial is considered a failure and aborts if a collision above a preset threshold occurs, or the vehicle deviates from the route by a preset margin.
Each trial is evaluated on six weathers, four of which are seen in training and two that are only used at test time.
The four training weathers are ``Clear noon'', ``Clear noon after rain'', ``Heavy raining noon'', and ``Clear sunset''.
The two test weathers are ``Wet sunset'' and ``Soft rain sunset''. 
We use CARLA 0.9.10 for all experiments.

\begin{table}[t]
\centering
\begin{tabular}{l@{\ \ }|c@{\ \ \ }c@{\ \ \ }c@{\ \ \ }|c@{\ }c@{\ }}
\toprule
Method & DS $\uparrow$ & RC $\uparrow$ & IS $\uparrow$ & Data & LiDAR \\ 
\midrule
CILRS~\cite{codevilla2019exploring} & $5.37$  & $14.40$ & $0.55$ & $-$ & $\times$ \\
LBC~\cite{chen2020learning}         & $8.97$  & $17.54$ & $\bf{0.73}$ & $-$ & $\times$ \\
Transfuser~\cite{prakash2021CVPR}   & $16.93$ & $51.82$ & $0.42$ & $150$K & $\checkmark$ \\
IA~\cite{toromanoff2020end}         & $24.98$ & $46.97$ & $0.52$ & $40$M   & $\times$ \\
\midrule
\textbf{\ourmethod} & $\bf{31.37}$ & $\bf{57.65}$ & $0.56$ & $1$M & $\times$ \\
\bottomrule
\end{tabular}
\caption{Comparison of the driving score (DS, main metric), route completion (RC), and infraction score (IS) on the CARLA leaderboard (accessed July 2021). For all three metrics, higher is better. Our method improves the driving score by $25\%$ relative to the prior state of the art~\cite{toromanoff2020end} while using $40\times$ less data.}
\lbltbl{leaderboard}
\end{table}

\begin{table}[t]
\centering
\begin{tabular}{l@{\ \ }c@{\ \ }c@{\ \ }c@{\ \ \ }c@{\ \ \ }c@{\ \ \ }c@{\ \ \ }}
\toprule
    Task && Town & Weather & IA & LBC & \textbf{\ourmethod} \\
\midrule
Empty &&\multirow{3}{*}{train} &\multirow{3}{*}{train} & $85$ & $89$ & $\mathbf{98}$ \\
Regular &&&& $85$ & $87$ & $\mathbf{100}$ \\
Dense &&&& $63$ & $75$ & $\mathbf{96}$ \\
\midrule
Empty &&\multirow{3}{*}{test} &\multirow{3}{*}{train} & $\it 77$ & $86$ & $\bf{94}$ \\
Regular &&&& $\it 66$ & $79$ & $\bf{89}$  \\
Dense &&&& $\it 33$ & $53$ & $\bf{74}$ \\
\midrule
Empty &&\multirow{3}{*}{train} &\multirow{3}{*}{test} & $-$ & $60$ & $\bf{90}$ \\
Regular &&&& $-$ & $60$ & $\bf{90}$ \\
Dense &&&& $-$ & $54$ & $\bf{84}$\\
\midrule
Empty &&\multirow{3}{*}{test} &\multirow{3}{*}{test} & $-$ & $36$ & $\bf{78}$\\
Regular &&&& $-$ & $36$ & $\bf{82}$ \\
Dense &&&& $-$ & $12$ & $\bf{66}$ \\
\bottomrule
\end{tabular}
\caption{Comparison of the success rate of the presented approach (\ourmethod) to the state of the art on NoCrash (LBC), and the winning entry of the 2020 CARLA Challenge (IA). All three methods are trained and evaluated on CARLA 0.9.10. IA uses all towns and all weathers to train. It thus does not have test weathers. {\it Italic} numbers indicate that the policy was trained on the test town. Additional route completion measurements are provided in the appendix for reference.}
\lbltbl{nocrash}
\end{table}

%CARLA leaderboard is a benchmark leaderboard that tests driving agents' under customized adversarial scenarios in unseen towns.
%Details of the metrics are described on the leaderboard website.

\paragraph{Comparison to the state of the art.}
\reftbl{leaderboard} compares the performance of the presented approach on the CARLA leaderboard. We list the three key metrics from the leaderboard: driving score (primary summary measure used for ranking entries on the leaderboard), route completion, and infraction score.
We compare to CILRS~\cite{codevilla2019exploring}, LBC~\cite{chen2020learning}, Transfuser~\cite{prakash2021CVPR} and IA~\cite{toromanoff2020end}.
LBC is the state of the art on the NoCrash benchmark, and Transfuser is a very recent method utilizing sensor fusion. Both LBC and Transfuser are based on imitation learning.
IA is the winning entry in the 2020 CARLA Challenge, and the prior leading entry on the CARLA leaderboard. IA is based on model-free reinforcement learning with Rainbow~\cite{hessel2018rainbow} and IQN~\cite{dabney2018implicit}.
% Our method is the $\#1$ entry on the 
% CARLA leaderboard at the time of writing.
In comparison to the prior leading entry (IA), we improve the driving score by $+25\%$ while using $40\times$ less data.

\reftbl{nocrash} compares the performance on the CARLA NoCrash benchmark.
We retrain LBC (the prior state of the art on NoCrash) on CARLA 0.9.10 using the same training data with augmented camera views as in our approach.
To help LBC generalize, we found it important to train with additional semantic segmentation supervision.
CARLA 0.9.10 features more complex visuals, and generalization to new weather conditions is harder.
IA features two models, a published model trained on CARLA 0.9.6 Town1 alone, and a much stronger CARLA Challenge model (trained on CARLA 0.9.10).
We compare to the stronger challenge model.
However, this model was trained on many more towns, and under both training and testing weather conditions.
It thus does not have held-out testing weathers.
Our method outperforms LBC and IA on all 12 tasks and conditions.
Furthermore, our method does not require expert actions anywhere in the training pipeline, unlike LBC.
We outperform IA on all traffic scenarios in both towns, even though we train only on Town1.

\begin{table}[t]
\centering
\begin{tabular}{l@{\ }c@{\ \ \ }|c@{\ \ \ }c@{\ \ \ }c@{\ \ \ }|c@{\ \ \ }}
\toprule
     \multicolumn{2}{l|}{Factorized world} & $\times$ & \checkmark & \checkmark & \checkmark \\
     Task & Town & DM & F-DM & CEM & \textbf{\ourmethod}  \\
\midrule
Straight & \multirow{2}{*}{train} & $\it 37$ & $\it 44$ & $\it \mathbf{100}$ & $\mathbf{100}$ \\
Turn && $\it 0$ & $\it 0$ & $\it 88$ & $\mathbf{100}$ \\
\midrule
Straight& \multirow{2}{*}{test} & $\it 44$ & $\it 52$ & $\it \mathbf{100}$ & $\mathbf{100}$ \\
Turn && $\it 0$ & $\it 0$ & $\it 97$ & $\mathbf{100}$ \\
\midrule
Empty & \multirow{3}{*}{train} & $0$ & $0$ & $\it 88$ & $\mathbf{98}$ \\
Regular && $\it 0$ & $\it 0$ & $\it 86$ & $\mathbf{100}$ \\
Dense && $\it 0$ & $\it 0$ & $\it 72$ & $\mathbf{96}$ \\
\midrule
Empty & \multirow{3}{*}{test} & $\it 0$ & $\it 0$ & $\it \bf 97$ & $94$ \\
Regular && $\it 0$ & $\it 0$ & $\it 84$ & $\bf{89}$\\
Dense && $\it 0$ & $\it 0$ & $\it 47$ & $\bf{74}$ \\

\bottomrule
\end{tabular}
\caption{Comparison of the success rate on the CoRL17 and NoCrash benchmark under training weathers. We compare our full visuomotor agent with model-based baselines. Dreamer (DM)~\cite{hafner2019dream} trains the full world model, whereas the rest follow our factorization and use the same forward model $\mathcal{T}^{ego}$ as our approach. Numbers in {\it italic} indicate agents that use privileged information (such as driving logs) at test time.
Our approach uses sensor readings alone. Nevertheless, our approach outperforms all baselines.}
\lbltbl{ablation}
\end{table}

\paragraph{Ablation study.}
\reftbl{ablation} compares our visuomotor agent with other model-based approaches.
All baselines optimize the same reward function described in \refsec{reward}.
Dreamer (DM)~\cite{hafner2019dream} trains a full-fledged embedding-based world model, and uses it to backpropagate analytic gradients to the policy during rollouts.
Building a full forward model of our driving scenarios can be challenging.
To help this baseline, we give it access to driving logs both during training and testing.
We additionally construct a variant, F-DM, which utilizes our factorized world model.
F-DM replaces a full embedding-based world model with our ego forward model $\mathcal{T}^{ego}$.
Akin to our method, it observes the pre-recorded world states and thus cannot backpropagate through a forward model of the world.
F-DM still trains the policy the same way as DM, using imaginary differentiable rollouts.
Since Dreamer is off-policy, we implement both DM and F-DM in an offline RL manner, and train both on the same dataset that is used to supervise our visuomotor agent.
CEM is an MPC baseline that factorizes the world and uses the cross-entropy method~\cite{mannor2003cross} to search for the best actions.
It uses our forward model, but cannot simulate the environment forward at the test time.
It assumes a static world.
Like Dreamer, CEM has access to the driving log at test time of the current timestep.
It replans at every timestep over the most recent driving log.
All the baselines use privileged information (driving logs), whereas our method takes sensor inputs alone.

We evaluate under the training weather for our method, as driving logs for baselines are weather-agnostic\footnote{The physics in CARLA does not vary with weather. Only sensor readings change with different weather conditions}.
We found that the NoCrash benchmark is too hard for the Dreamer baseline, and thus additionally test on the much easier CoRL17 benchmark~\cite{dosovitskiy17}.
Akin to NoCrash, each task in the CoRL17 benchmark contains 50 predefined routes: 25 for the training town and 25 for an unseen test town.
It runs on empty roads with simpler routes compared to NoCrash.
Our method outperforms all other model-based baselines on almost all tasks by a margin, despite using sensor inputs instead of driving logs.
Dreamer with a factorized world model outperforms the full world model but still fails to generalize beyond straight driving.
One reason for the poor performance of Dreamer may be a bias in the training set.
Cars mostly drive straight.
Dreamer may simply see too few turning scenarios compared to the endless stream of straight driving.

\begin{figure*}[t]
    \begin{subfigure}[b]{0.225\linewidth}
     \centering
     \includegraphics[height=0.5\textwidth]{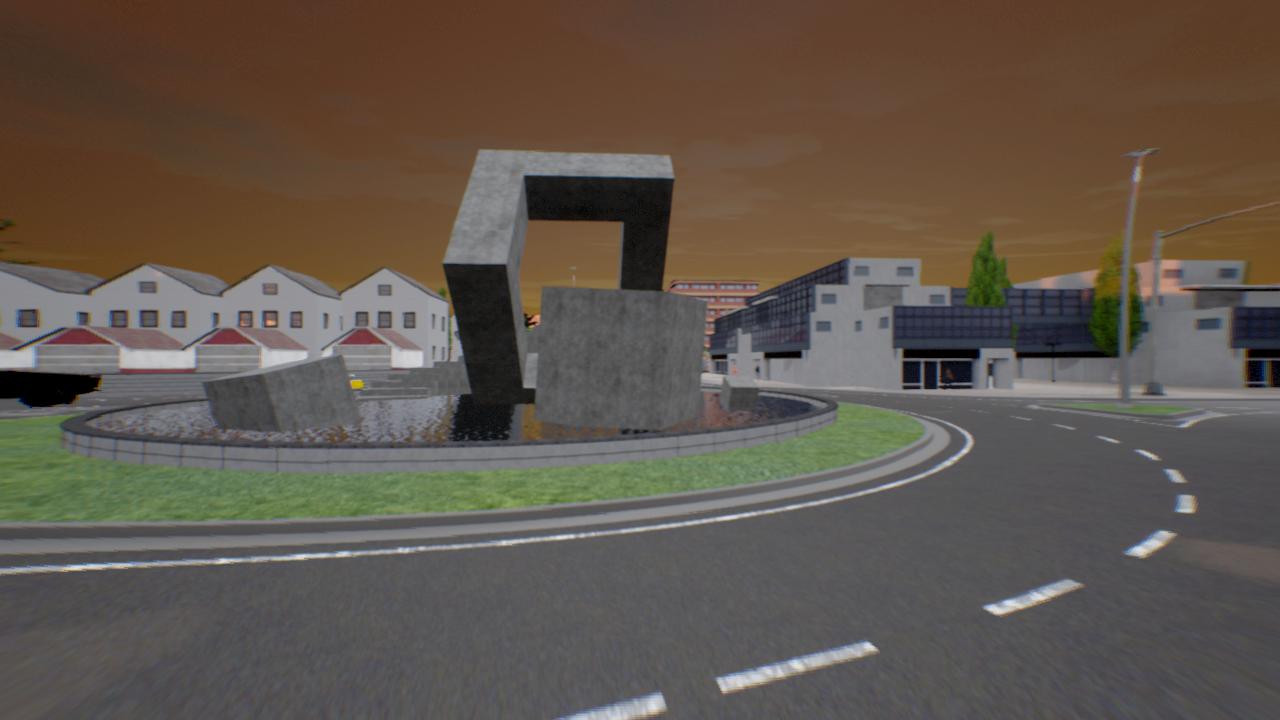}\vspace{1ex}
     \includegraphics[height=0.5\textwidth]{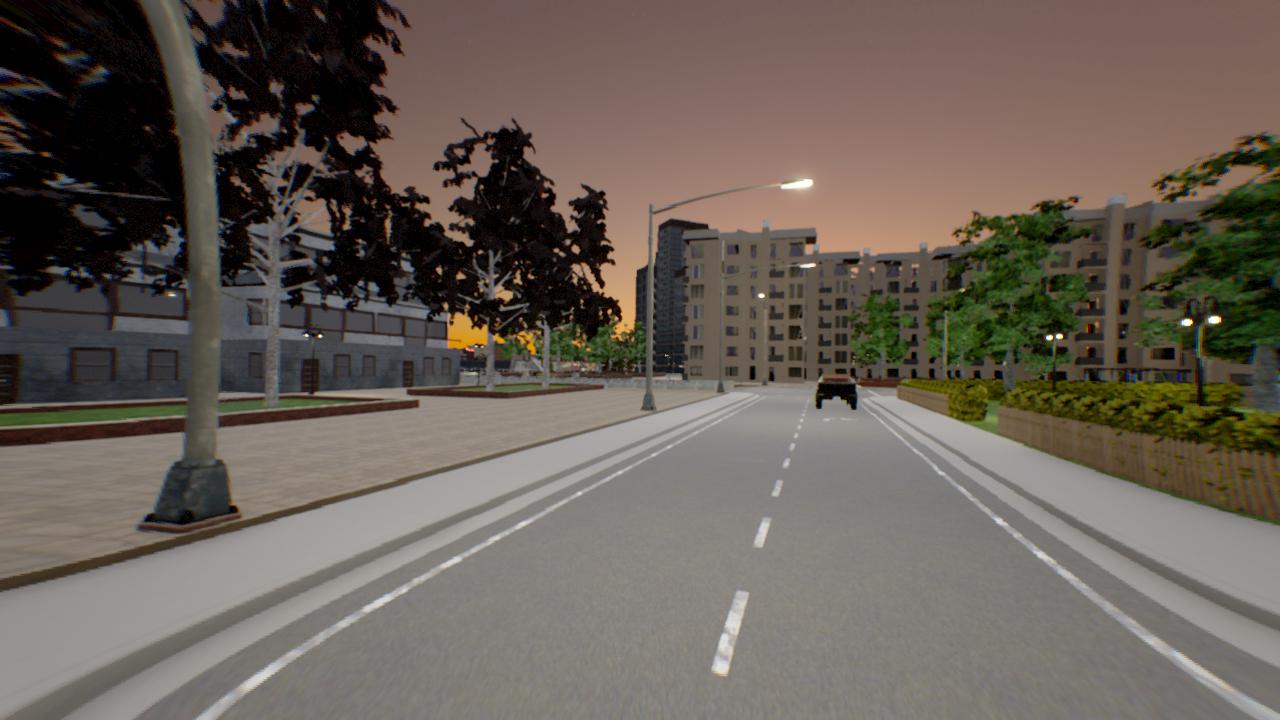}\vspace{1ex}
     \includegraphics[height=0.5\textwidth]{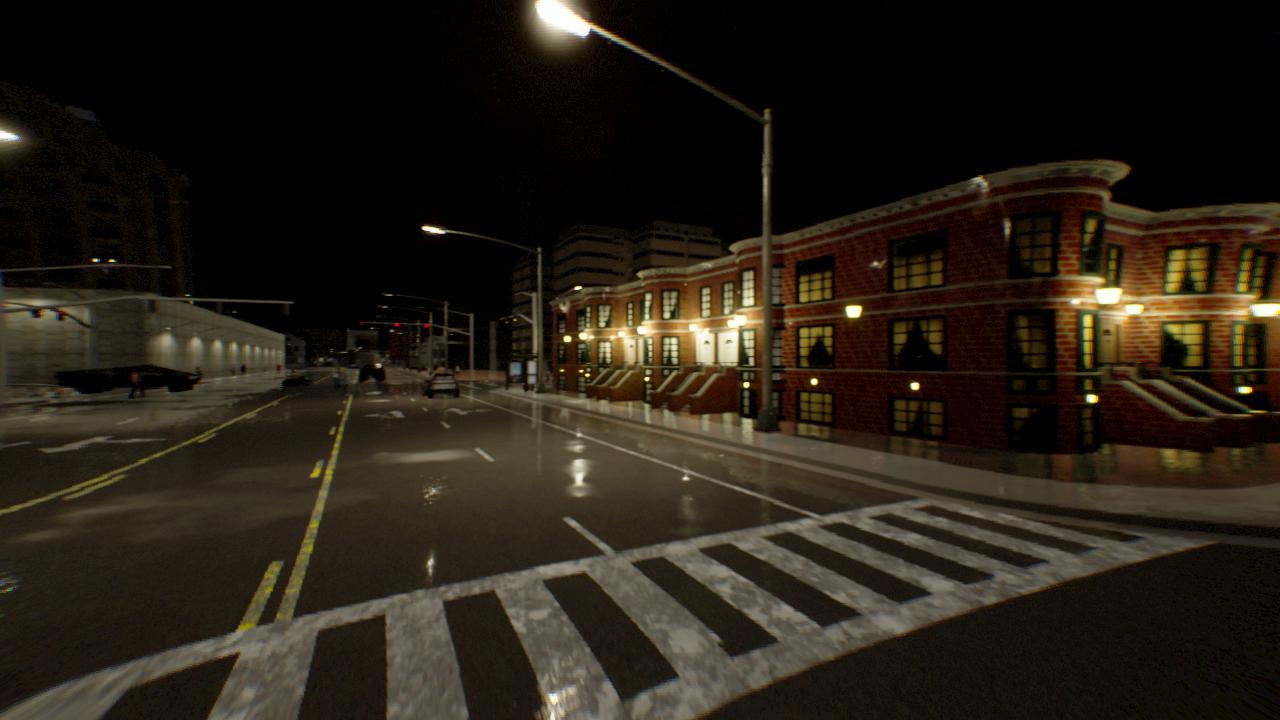}\vspace{1ex}
     \includegraphics[height=0.5\textwidth]{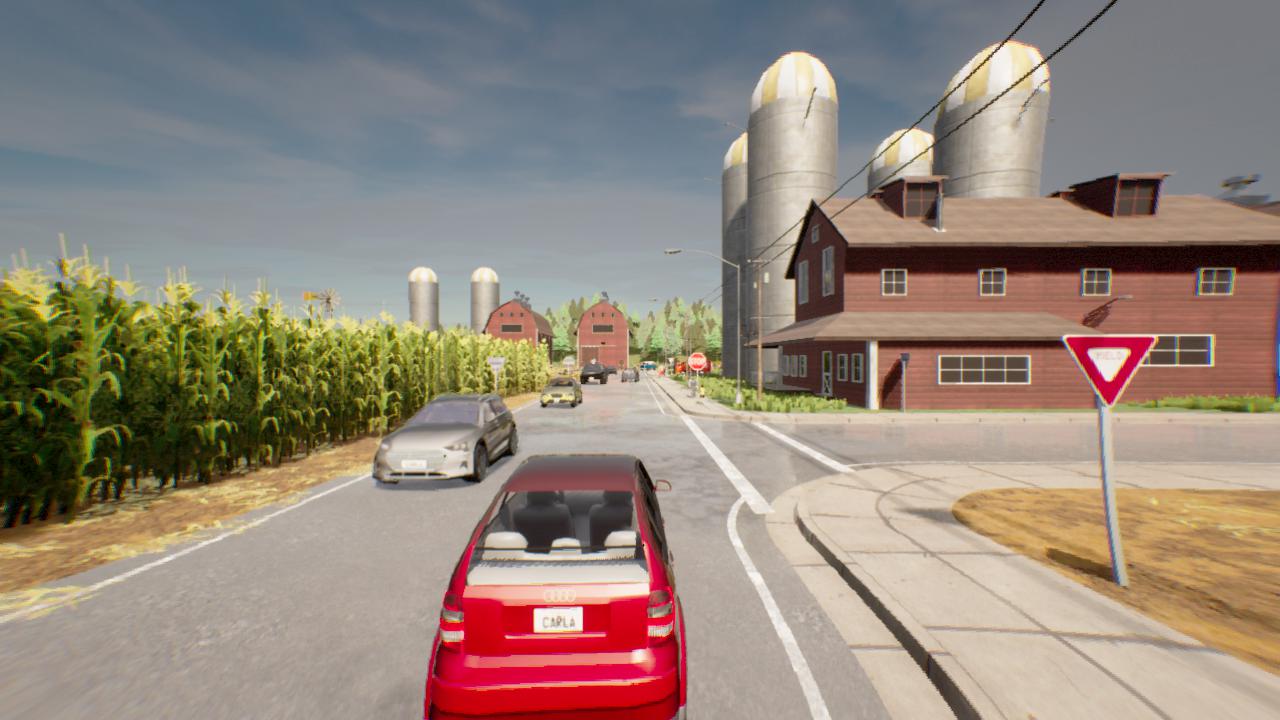}
    \caption{RGB camera}
    \end{subfigure}%
    \begin{subfigure}[b]{0.225\linewidth}
     \centering
     \includegraphics[height=0.5\textwidth]{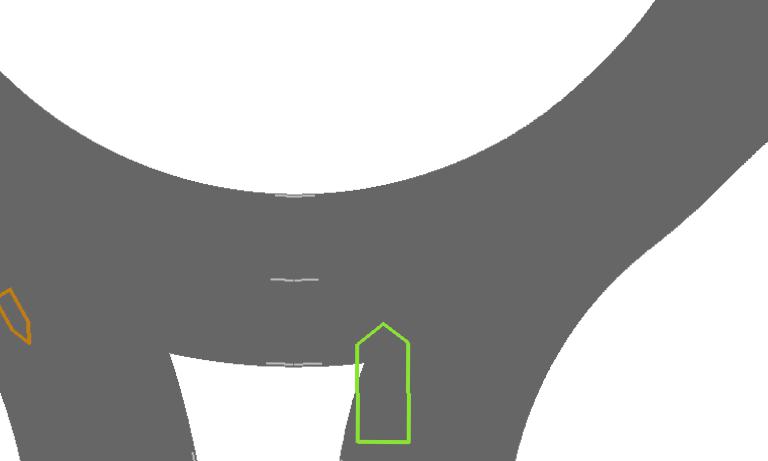}\vspace{1ex}
     \includegraphics[height=0.5\textwidth]{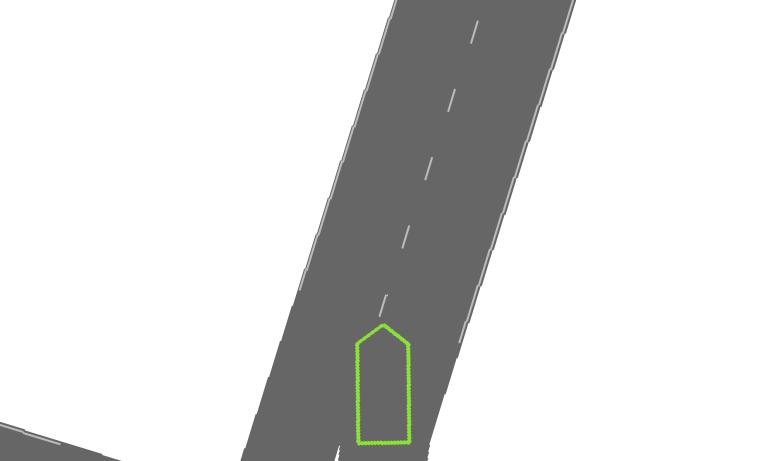}\vspace{1ex}
     \includegraphics[height=0.5\textwidth]{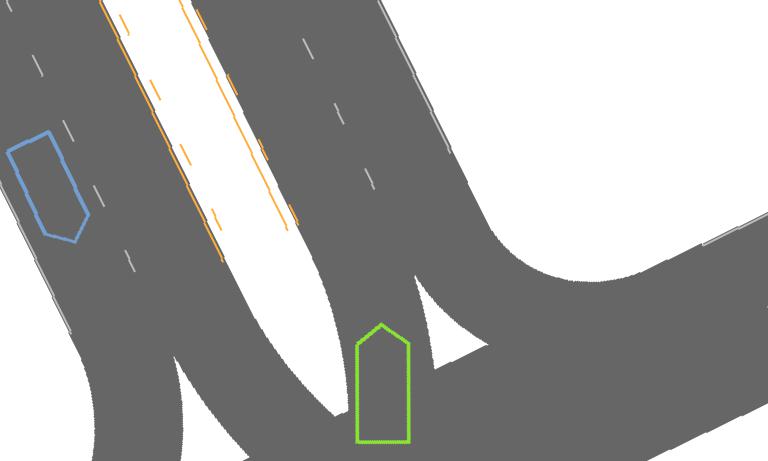}\vspace{1ex}
     \includegraphics[height=0.5\textwidth]{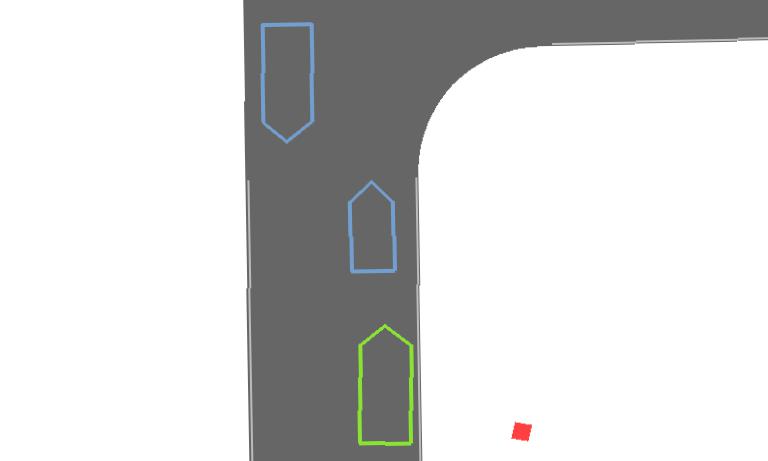}
    \caption{Map}
    \end{subfigure}%
    \begin{subfigure}[b]{0.225\linewidth}
     \centering
     \includegraphics[height=0.5\linewidth]{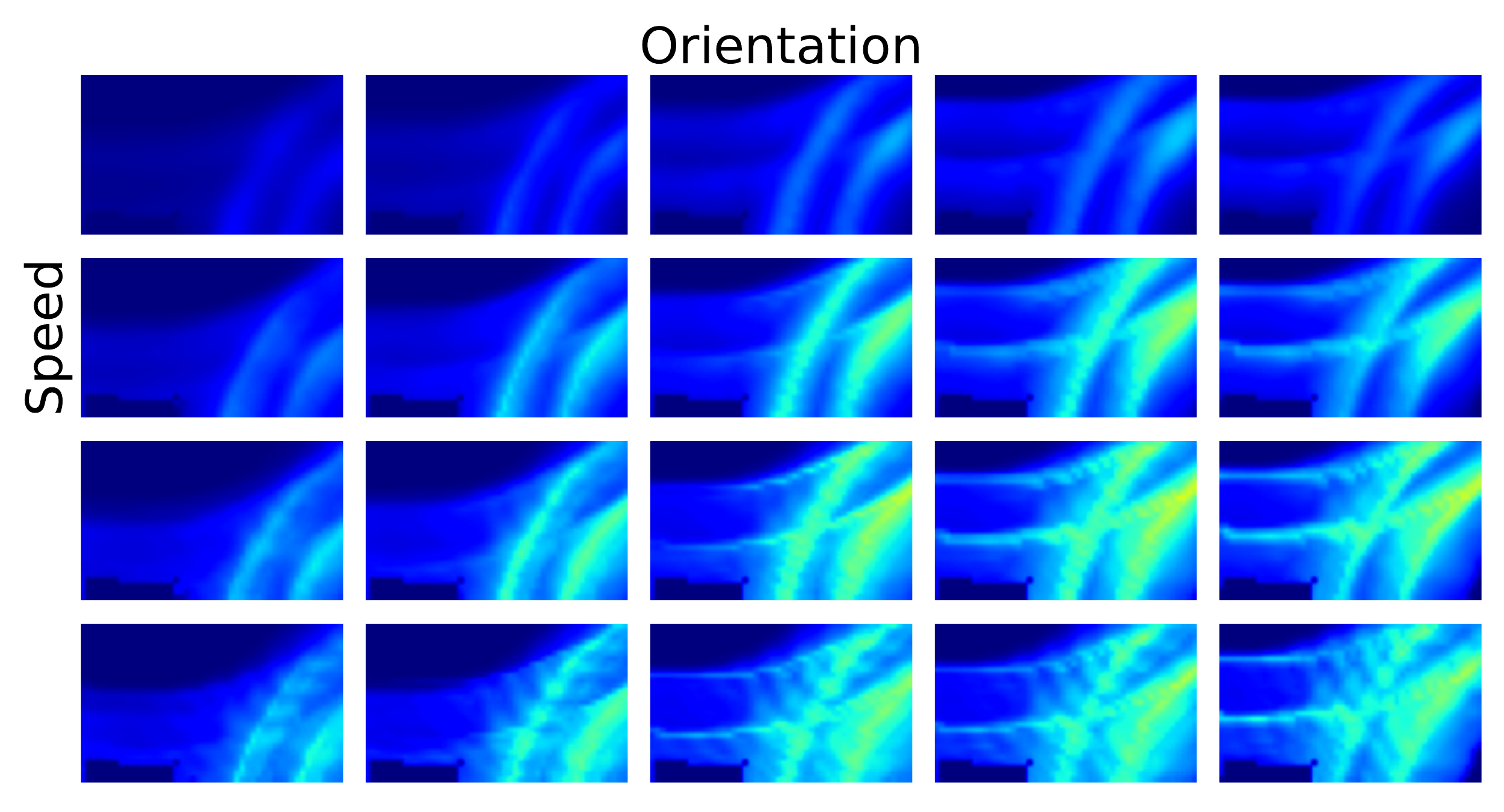}\vspace{1ex}
     \includegraphics[height=0.5\linewidth]{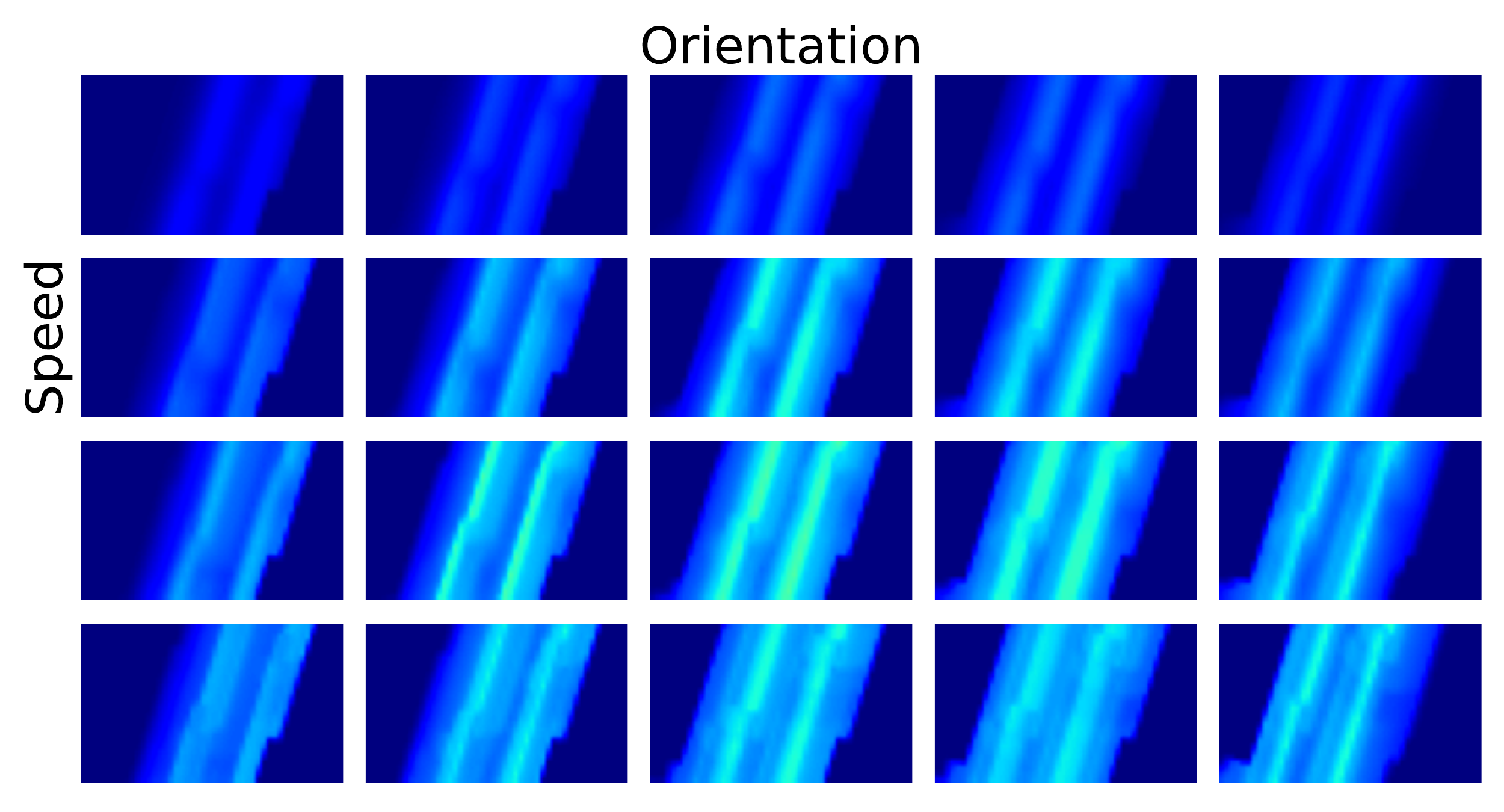}\vspace{1ex}
     \includegraphics[height=0.5\linewidth]{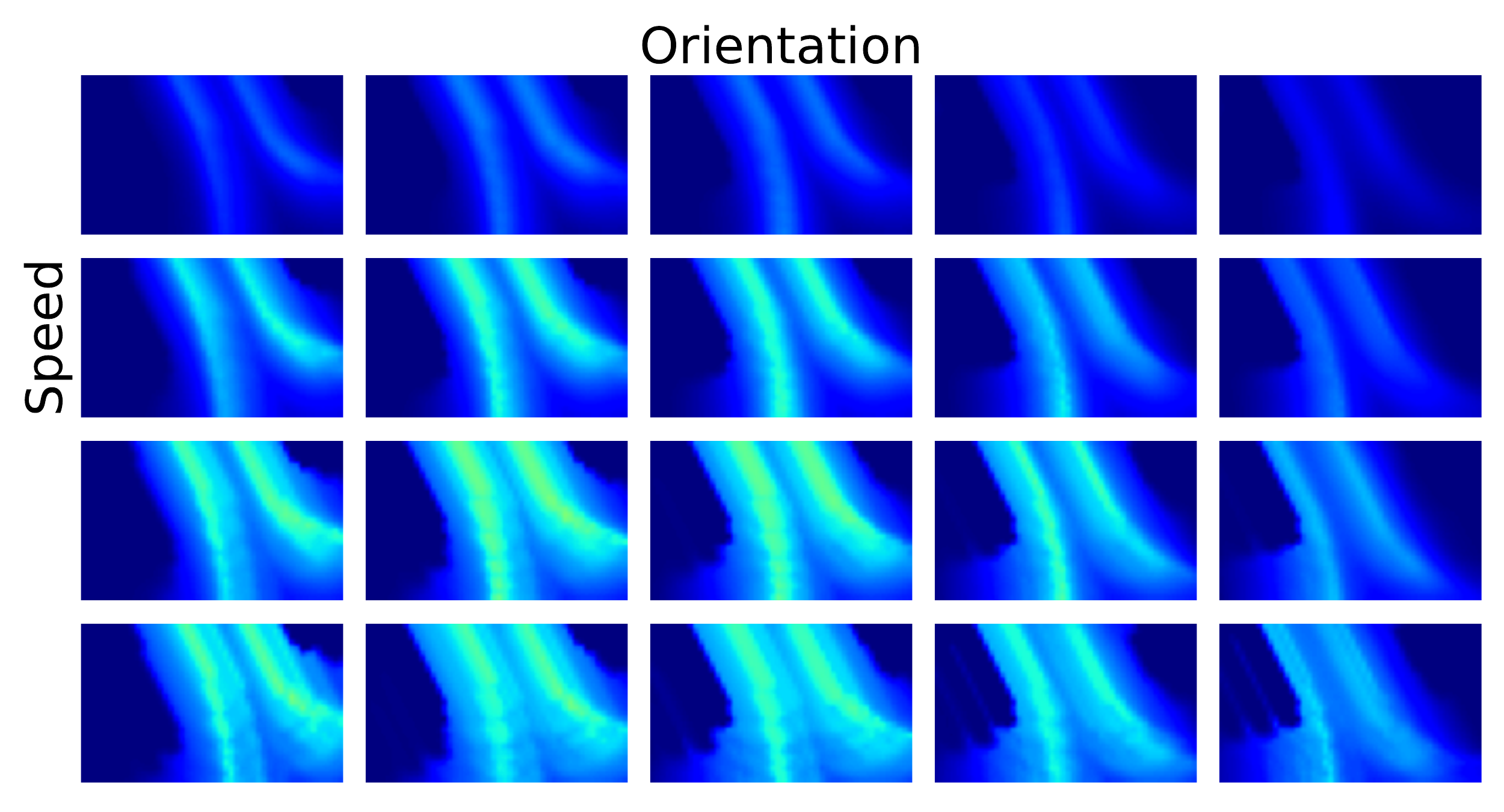}\vspace{1ex}
     \includegraphics[height=0.5\linewidth]{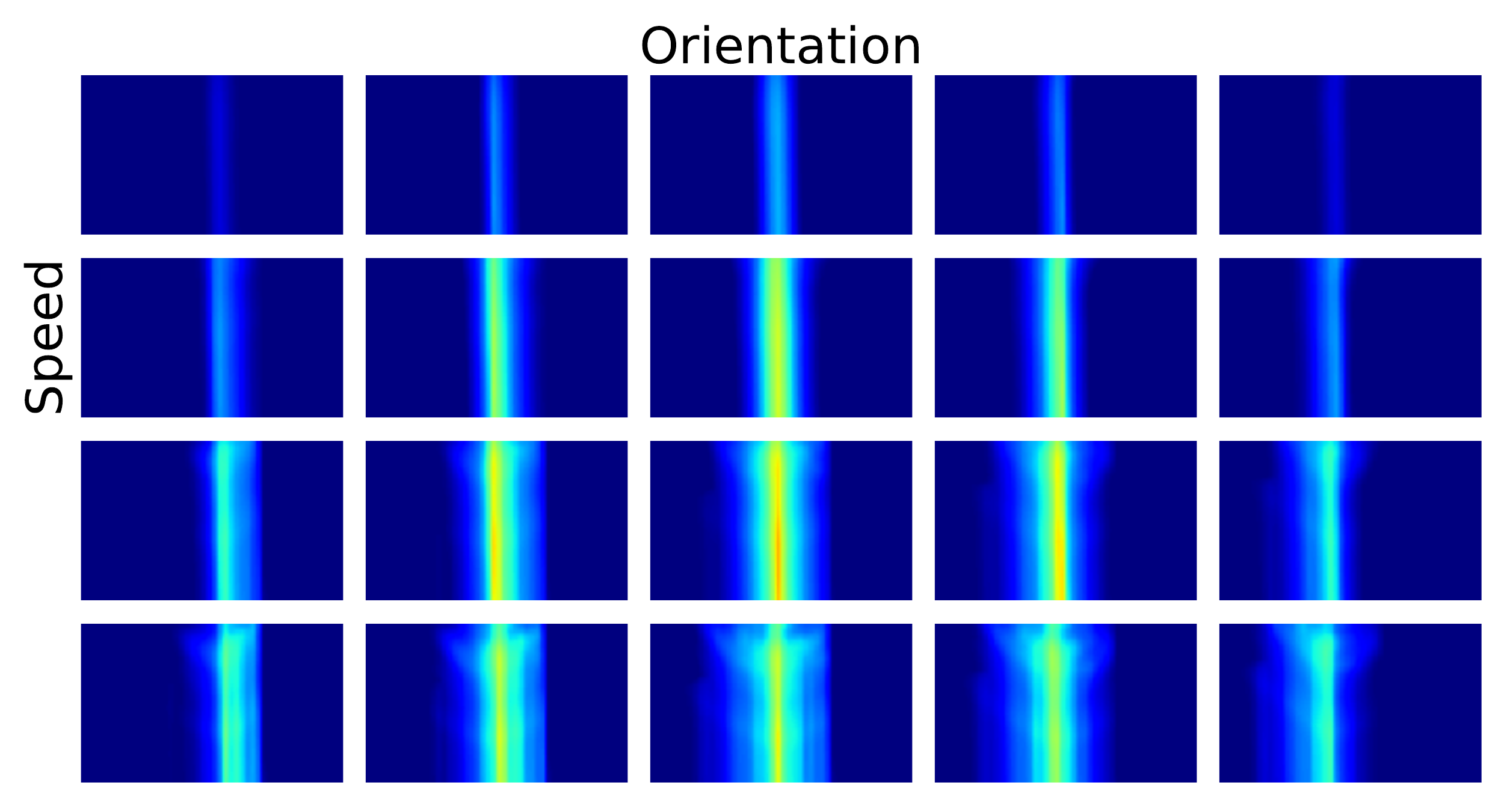}
    \caption{Value map}
    \end{subfigure}%
    \begin{subfigure}[b]{0.225\linewidth}
     \centering
     \includegraphics[height=0.5\linewidth]{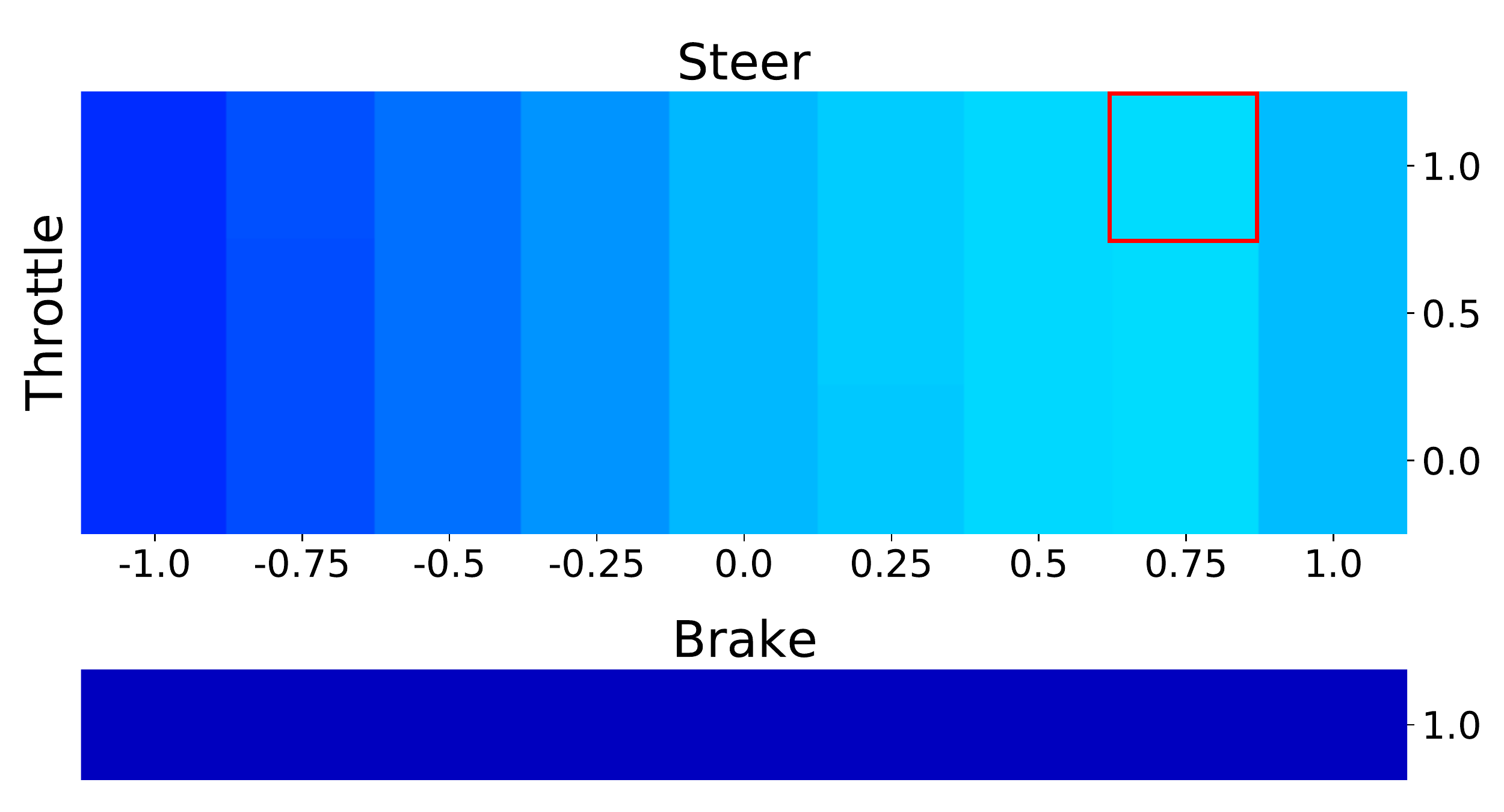}\vspace{1ex}
     \includegraphics[height=0.5\linewidth]{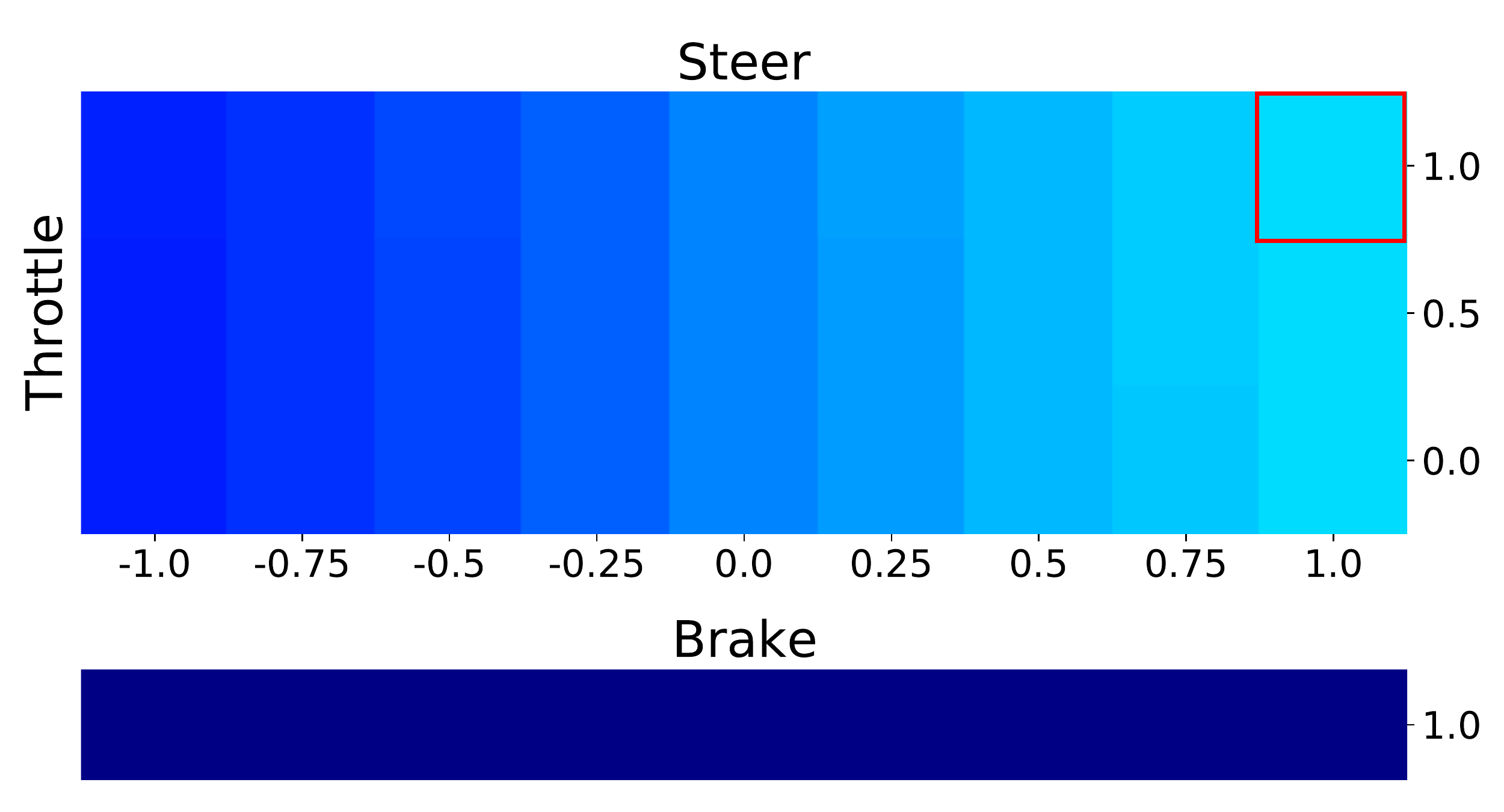}\vspace{1ex}
     \includegraphics[height=0.5\linewidth]{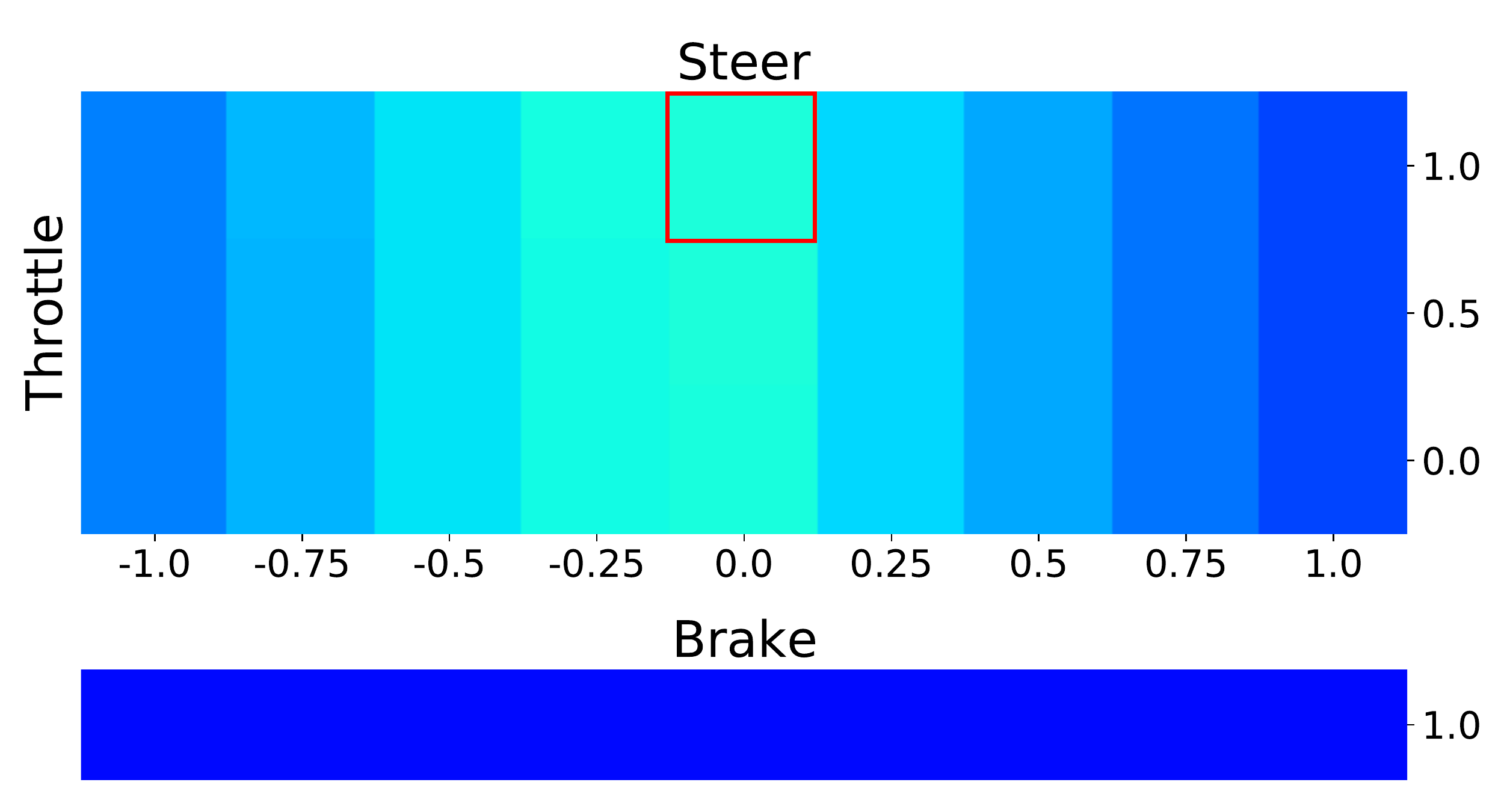}\vspace{1ex}
     \includegraphics[height=0.5\linewidth]{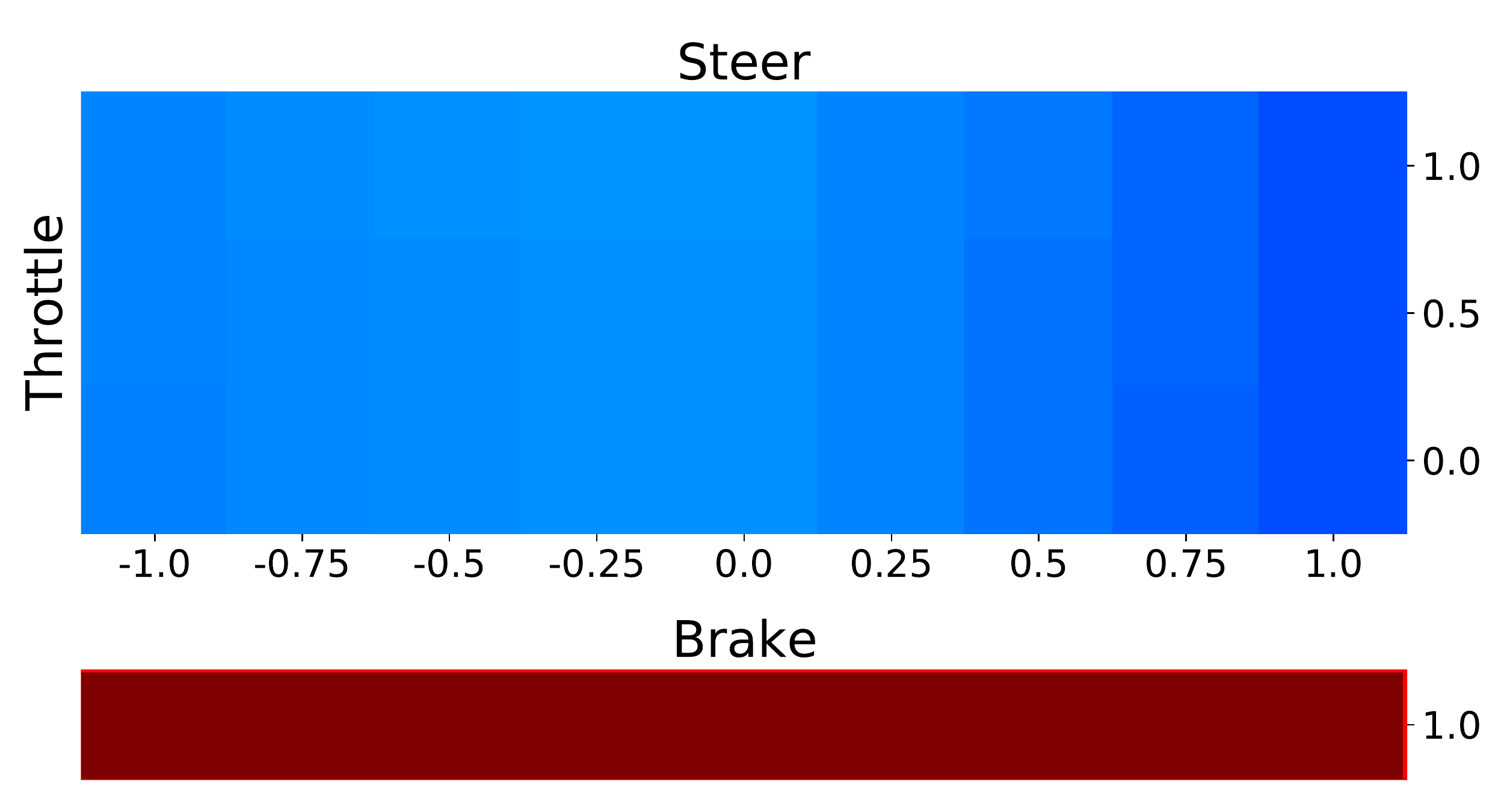}
    \caption{Action-value}
    \end{subfigure}%
    \begin{subfigure}[b]{0.1\linewidth}
     \centering
     \includegraphics[height=4.75\linewidth]{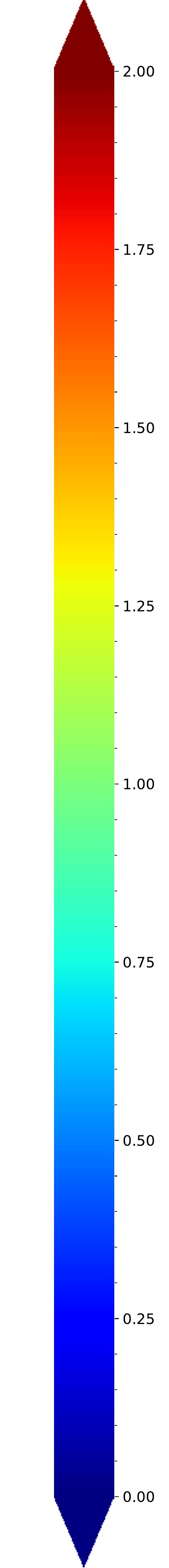}
     \caption*{}
    \end{subfigure}%
     \caption{Visualization of the computed value function and action-value function for the current frame. The RGB camera image (a) and bird-eye view maps (b) show the ego-vehicle location in the world. The value-maps (c) show the discretized tabular value estimate for 4 speed bins $\times$ 5 orientation bins. The orientation bins are $-95^\circ$ to $95^\circ$ from left to right, and the speed bins are $0$ m/s to $8$ m/s from top to bottom. Each map has a resolution of $96 \times 96$ corresponding to a $24$m$^2$ area around the vehicle. We crop areas behind the ego-vehicle for visualization. The value maps use 5 Bellman updates and see 1.25s into the future. (d) shows the action-values based on the current ego-vehicle state. Actions with highest values are highlighted with red boxes. These action-values supervise the visuomotor policy that takes camera RGB images as input.}
\lblfig{visualization}
\end{figure*}

\begin{table*}[t]
\centering
\begin{tabular}{c@{\ }c|c@{\ \ }c@{\ \ }c|c@{\ \ }c@{\ \ }c|c@{\ \ }c@{\ \ }c|c@{\ \ }c@{\ \ }c}
\toprule
    & & \multicolumn{6}{c|}{Train town} & \multicolumn{6}{c}{Test Town} \\
    & & \multicolumn{3}{c|}{Train Weather} & \multicolumn{3}{c|}{Test Weather} & \multicolumn{3}{c|}{Train Weather} & \multicolumn{3}{c}{Test Weather} \\
    CA & SA & Empty & Regular & Dense & Empty & Regular & Dense & Empty & Regular & Dense & Empty & Regular & Dense \\
\midrule
    $\times$ & $\times$ &             $87$ & $82$ & $82$ & $60$ & $78$ & $82$ & $85$ & $80$ & $63$ & $68$ & $54$ & $42$ \\
    $\times$ & $\checkmark$&           $97$ & $97$ & $92$ & $78$ & $82$ & $80$ & $92$ & $\bf{91}$ & $64$ & $66$ & $72$ & $58$ \\
    $\checkmark$& $\times$ &      $\bf{100}$& $98$ & $90$ & $\bf{92}$ & $\bf{94}$ & $76$ & $90$ & $82$ & $60$ & $78$ & $62$ & $48$ \\
    $\checkmark$&$\checkmark$    & $98$ & $\bf{100}$& $\bf{96}$ & $90$ & $90$ & $\bf{84}$ & $\bf{94}$ & $89$ & $\bf{74}$ & $\bf{78}$ & $\bf{82}$ & $\bf{66}$ \\
\bottomrule
\end{tabular}
\caption{Comparison of success rate in the NoCrash benchmark under different ablation conditions. \textbf{CA} stands for ``camera augmentation'' and \textbf{SA} stands for ``speed augmentation''. All ablation models are trained on the same dataset and evaluated on CARLA 0.9.10. \textbf{CA} models additionally train on two augmented camera views per dataset frame.}
\lbltbl{visuomotor_option}
\end{table*}

\begin{table}[t]
\centering
\begin{tabular}{c@{\ \ \ }c@{\ \ \ }c@{\ \ \ }c@{\ \ \ }}
\toprule
    % \multicolumn{2}{l|}{Segmentation auxilliary loss} & $-$ & \checkmark \\
    && \multicolumn{2}{c}{Auxilliary loss} \\
    Town & Weather & $\times$ & \checkmark \\
\midrule
   train & train & $95$ & $100$ \\
   train & test & $70$ & $100$ \\
   test & train & $80$ & $98$ \\
   test & test & $46$ & $76$ \\
\bottomrule
\end{tabular}
\caption{Comparison of success rate in the NoCrash benchmark on the empty traffic condition with and without the auxiliary semantic segmentation loss.}
\lbltbl{seg_aux}
\vspace{-1em}
\end{table}
\reftbl{visuomotor_option} compares different variation of our visuomotor agent at the distillation stage. \textbf{CA} stands for camera augmentation, meaning the model trains on the additional augmented camera images, described in \refsec{dataset}. \textbf{SA} stands for ``speed augmentation''. An \textbf{SA} model trains to predict action values on all discretized speed bins, instead of taking as input the recorded speed reading from the dataset. During test time, an \textbf{SA} models uses linear interpolation to extract the action-values corresponding to the ego-vehicle speed. Models trained with camera or speed augmentation consistently outperform ones that were not, showing the benefits of dense action-values computed using our factorized Bellman updates.
We therefore use camera and speed augmentation for our models for the CARLA leaderboard and the NoCrash benchmark.
With the augmented supervision extracted from the dense action-values, models perform well even without techniques such as trajectory noise injection~\cite{laskey2017dart,codevilla2019exploring}. Results for models trained with injected steering noise are provided in the appendix for reference.

\reftbl{seg_aux} compares our visuomotor agent, which is trained with an auxiliary semantic segmentation loss, with a simpler baseline that does not use this auxiliary loss.
Policies trained with semantic segmentation consistently outperform the action-only baseline, especially under generalization settings.
We observed the same for the LBC baseline, which also uses semantic segmentation as an auxiliary loss.

\paragraph{Traffic light infraction analysis.}
We additionally analyze traffic light infractions on the NoCrash benchmark.
\reftbl{traffic} compares the average number of traffic light violations per hour on all trials in the NoCrash benchmark.
The presented approach has fewer traffic light infractions than the reinforcement learning baseline (IA) on all six tasks under the training weathers.

\paragraph{Visualization.}
\reffig{visualization} shows a visualization of the computed value and action-value functions for various driving scenarios.
Each of these action-value functions densely supervises the policy for the displayed image.

\begin{table}[b]
\centering
\begin{tabular}{l@{\ \ \ \ \ }c@{\ \ \ \ \ }c@{\ \ \ \ \ }c@{\ \ \ \ \ }c@{\ \ \ \ \ }c@{\ \ \ \ \ }}
\toprule
\multicolumn{3}{l}{Oracle actions} & $\times$ & $\checkmark$ & $\times$ \\
Task & Town & Weather & IA & LBC & \textbf{\ourmethod}  \\
\midrule
Empty & \multirow{3}{*}{train} & \multirow{3}{*}{train} & $3.34$ & $1.35$ & $\bf{0.00}$ \\
Regular &&& $6.71$ & $1.89$ & $\bf{0.43}$  \\
Dense &&& $15.41$ & $3.27$ & $\bf{2.61}$ \\
\midrule
Empty & \multirow{3}{*}{test} & \multirow{3}{*}{train} & $62.18$ & $\bf{8.45}$ & $10.68$ \\
Regular &&& $53.28$ & $8.22$ & $\bf{6.95}$ \\
Dense &&& $54.94$ & $\bf{7.26}$ & $12.90$ \\
\midrule
Empty & \multirow{3}{*}{train} & \multirow{3}{*}{test} & $-$ & $0.36$ & $\bf{0.00}$  \\
Regular &&& $-$ & $0.81$ & $\bf{0.00}$ \\
Dense &&& $-$ & $\bf{0.52}$ & $4.29$\\
\midrule
Empty & \multirow{3}{*}{test} & \multirow{3}{*}{test} & $-$ & $\bf{8.17}$ & $14.46$ \\
Regular &&& $-$ & $\bf{8.61}$ & $11.30$ \\
Dense &&& $-$ & $\bf{4.87}$ & $13.28$ \\
\bottomrule
\end{tabular}
\caption{Comparison of the average number of traffic light violations per hour of trials on the NoCrash benchmark.
We compare our approach to LBC (prior state of the art on NoCrash) and IA (the winning entry of the 2020 CARLA challenge).
LBC trains from oracle trajectories, whereas IA and ours do not.}
\lbltbl{traffic}
\vspace{-1em}
\end{table}

\begin{figure*}[t]
\centering
\begin{subfigure}[b]{0.25\linewidth}\includegraphics[width=\linewidth]{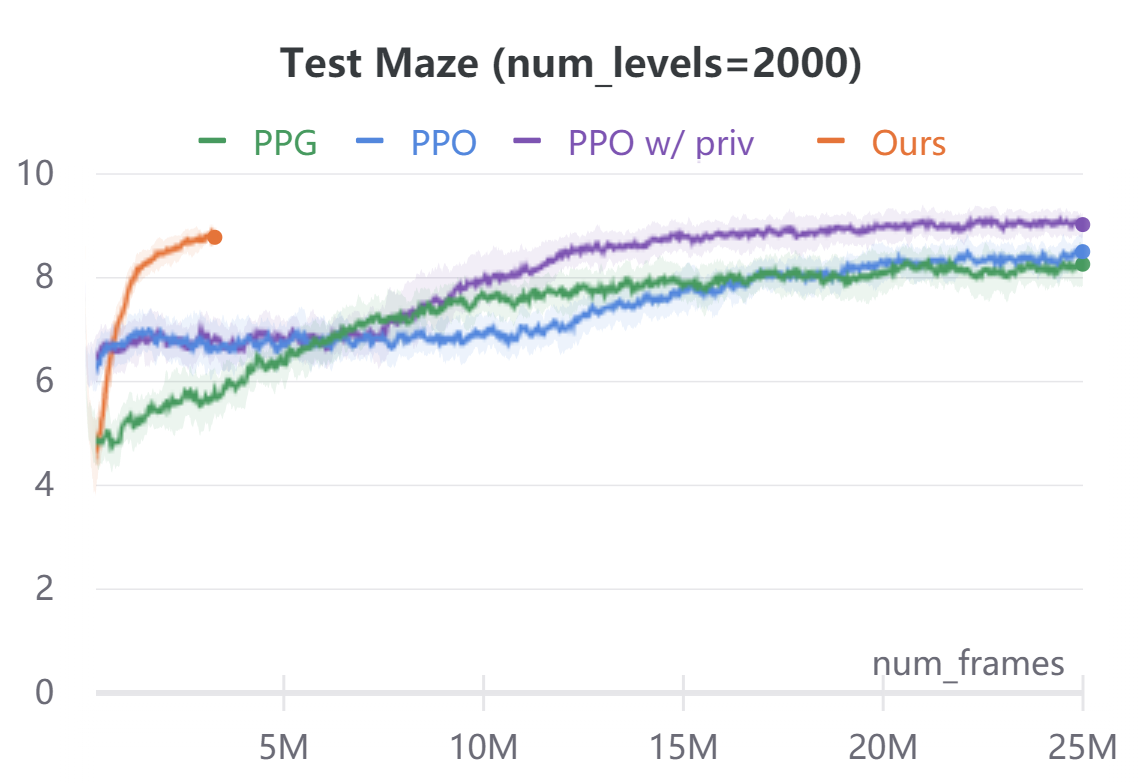}\caption{Maze 2000 training levels}\end{subfigure}%
\begin{subfigure}[b]{0.25\linewidth}\includegraphics[width=\linewidth]{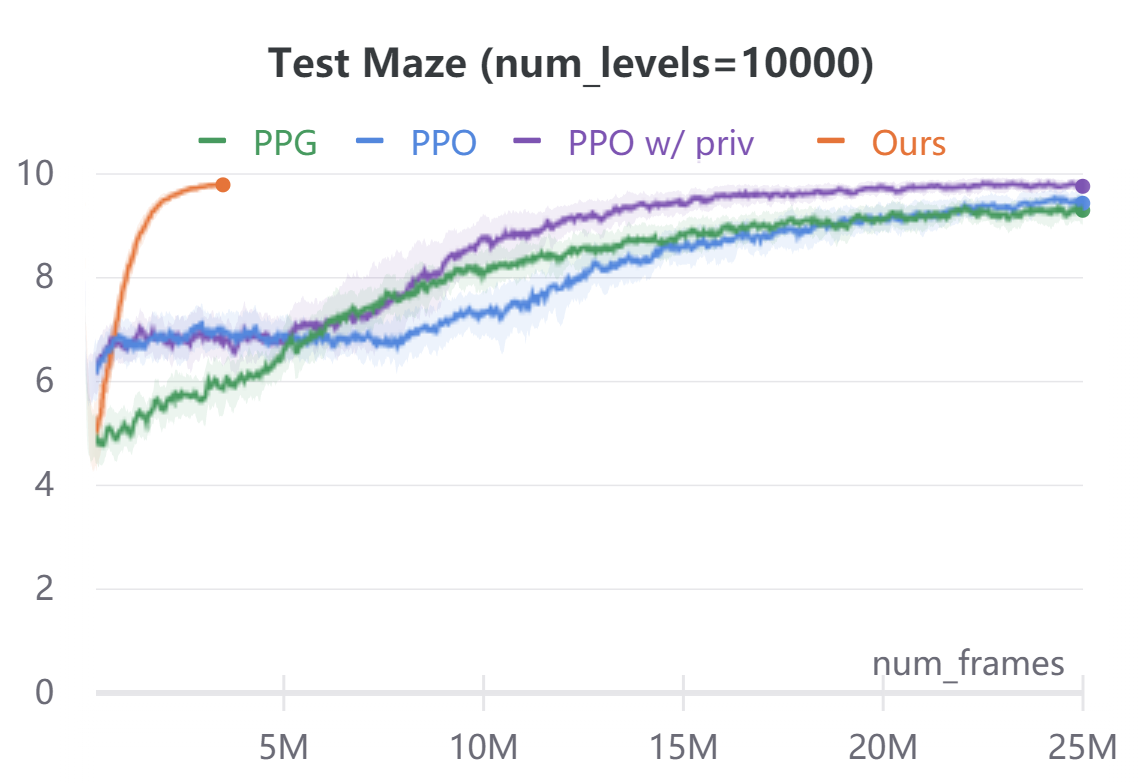}\caption{Maze 10000 training levels}\end{subfigure}%
\begin{subfigure}[b]{0.25\linewidth}\includegraphics[width=\linewidth]{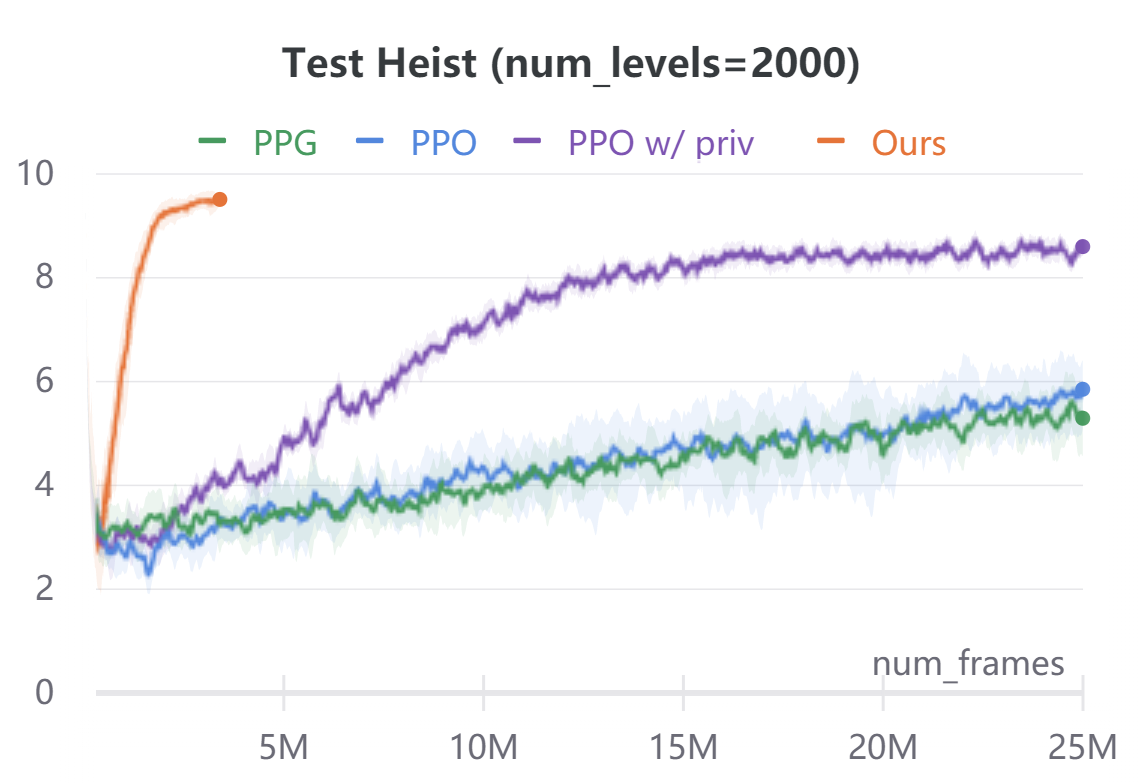}\caption{Heist 2000 training levels}\end{subfigure}%
\begin{subfigure}[b]{0.25\linewidth}\includegraphics[width=\linewidth]{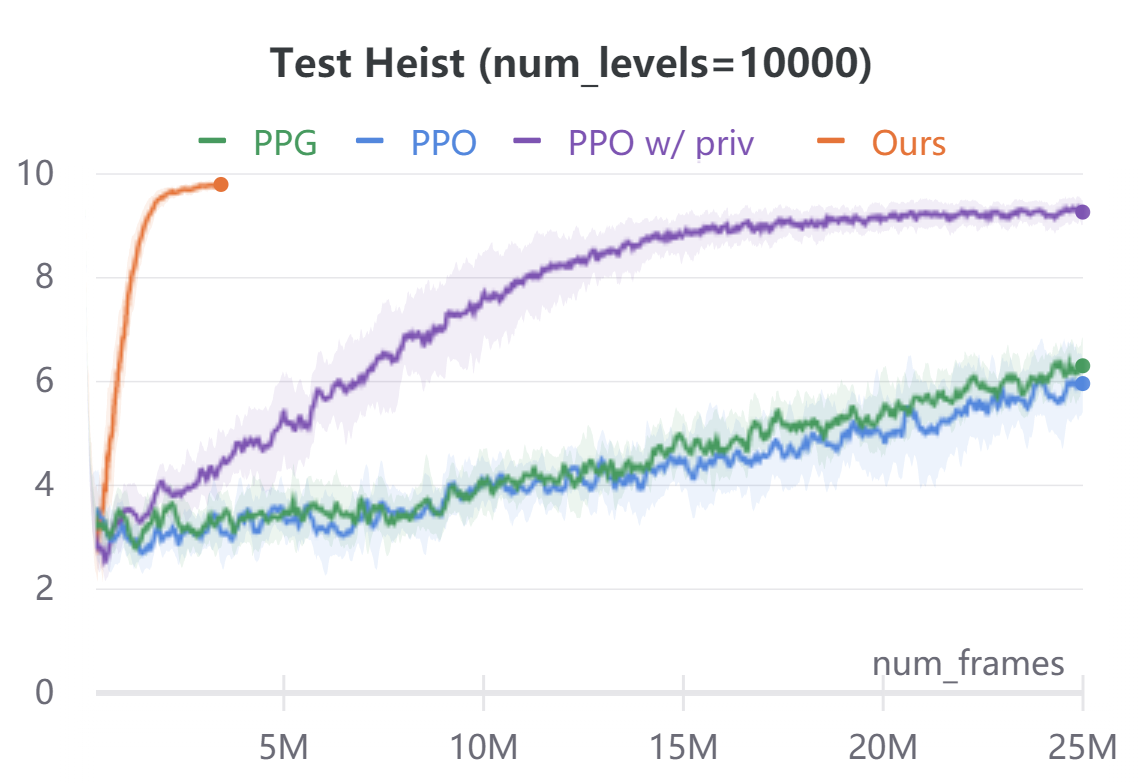}\caption{Heist 10000 training levels}\end{subfigure}%

\caption{Comparison of our method to state-of-the-art model-free reinforcement learning on the navigational tasks in the ProcGen benchmark.
All plots measure the average episode returns on the testing levels.
`PPO w/ priv' is a customized PPO implementation that during training additionally takes as input the same privileged information that our approach uses to compute rewards and train the agent forward model. The presented approach is an order of magnitude more sample-efficient.}
\lblfig{procgen_test}
\vspace{-1em}
\end{figure*}

\paragraph{ProcGen navigation.}
To demonstrate the broad applicability of our approach, we additionally evaluate on the navigational tasks (\textit{Maze} and \textit{Heist}) in the ProcGen benchmark~\cite{cobbe2019leveraging}.
In both environments, the agent is rewarded for navigating to desired locations.
\textit{Maze} features a plain navigation task through a complex environment.
\textit{Heist} additionally requires the agent to collect keys and unlock the doors before navigating to the goal.
In ProcGen, the action space is discrete, hence we only discretize the ego-agent's states.
Similar to CARLA, we discretize the agent state into $N_H\times N_W$ location bins and $N_\theta$ orientation bins.
We use $N_H=N_W=32$, and $N_\theta=8$.
We ignore velocity.
The agent's forward dynamics model in ProcGen is not location agnostic as in CARLA.
To address this, we use a small ConvNet to extract the environment context around the ego-agent forward model $\mathcal{T}^{ego}$.
The ConvNet takes as input a cropped $13\times13$ region around the ego-agent in the original $64\times64$ RGB observations. The ConvNet features are concatenated with agent orientation to predict the next ego-agent's states under all discrete action commands.
In order to evaluate sample efficiency, we implement our method on ProcGen in an off-policy reinforcement learning manner.
We alternate between training or fine-tuning a policy and forward model, and rolling out new trajectories under the current policy.
Compared to model-free baselines, our approach needs access to a dense reward function, instead of just the scalar reward signal of the environment.
We compute this reward function using semantic labels obtained via the ProcGen renderer.
For \textit{Maze}, the reward function awards $+1$ for goal location regardless of orientation. For \textit{Heist}, the reward function awards $+1$ for key and unlockable door locations regardless of orientation. In addition, we mask all unachievable ego-state values to $0$ during the Bellman equation evaluation.
We use this privileged information in the action-value computation only, and in no other place in our algorithm.
\reffig{procgen_test} compares the performance and sample-efficiency of our method with model-free reinforcement learning baselines PPO~\cite{schulman2017proximal} and PPG~\cite{cobbe2020ppg}.
PPG is the current state of the art on the ProcGen benchmark.
In addition, we compare to a customized PPO implementation which during training also takes as input the same privileged information used in our method.
Our method converges within $3$M frames, while model-free baselines take up to $25$M frames.
For both \textit{Maze} and \textit{Heist} environments, we train all agents on two different conditions: $2000$ and $10000$ (procedurally generated) training levels.
For both environments, agents are tested on completely randomized procedurally-generated levels.
The comparison of average episode returns on the training levels is in the appendix for reference.
Our method is an order of magnitude more sample-efficient than all the model-free RL baselines even when those methods are given the same privileged information used by our reward computation.

\section{Conclusion}
We show that assuming independence between the agent and the environment, which we refer to as a \emph{world on rails}, significantly simplifies modern reinforcement learning.
While true independence rarely holds, the gains in training efficacy outweigh the modeling constraints.
Even with a simple reward function, an agent trained in a world-on-rails learns to drive better than state-of-the-art imitation learning agents on standard driving benchmarks.
In addition, the presented policy learning framework is an order of magnitude more sample-efficient than state-of-the-art reinforcement learning on challenging ProcGen navigation tasks.

\section*{Acknowledgements}
We thank Yuke Zhu for his valuable feedback.
We thank Tianwei Yin for his help on figure 1.
This work was supported by the NSF Institute for Foundations of Machine Learning and NSF award \#1845485.

{\small
\bibliographystyle{ieee_fullname}
\bibliography{egbib}
}

%\begin{appendices}
\appendix

\section*{\Large Appendix}

\section{Kinematic Bicyle Model}
The kinematics of the bicycle model~\cite{polack2017} $\mathcal{T}^{ego}$ used in our CARLA experiment is described below: 
\begin{align*}
    \dot{x} &= v \cos(\theta+\beta)\lbleq{bicycle} \\
    \dot{y} &= v \sin(\theta+\beta)\notag \\
    \dot{v} &= a\notag \\
    \dot{\theta} &= \frac{v}{r_b}\sin(\beta)\notag \\
    \tan(\beta) &= \frac{r_b}{f_b+r_b}\tan(\phi)\notag
\end{align*}
We train $\mathcal{T}^{ego}$ in an auto-regressive manner using L1 loss and stochastic gradient descent:
\begin{align*}
    J_{ego}&=\sum_{t}^T|x_t-\hat{x}_t|+|y_t-\hat{y}_t|\\
    &+\sum_{t}^T|\cos(\theta_t)-\cos(\hat{\theta}_t)|+|\sin(\theta_t)-\sin(\hat{\theta}_t)|
\end{align*}
where $x_{t+1},y_{t+1},\theta_{t+1},v_{t+1}=\mathcal{T}^{ego}(x_{t},y_{t},\theta_{t},v_{t}, a_t)$, and $a_t = (s_t, t_t, b_t)$. We only model $\theta$ as a transform of $s$; $a$ as a transform of $(t,b)$, and vehicle wheelbases $r_b, f_b$. 
We use an action repeat of 5 frames, hence both data collection and planning operate at 4 FPS, whereas the simulator and the visuomotor policy run at 20 FPS. 

\section{ProcGen Training Levels Returns}
\reffig{procgen_train} plots the average episode returns of our method against PPO~\cite{schulman2017proximal}, PPG~\cite{cobbe2020ppg}, and PPO with access to privileged information.

\begin{figure*}[t]
\centering
\begin{subfigure}[b]{0.25\linewidth}\includegraphics[width=\linewidth]{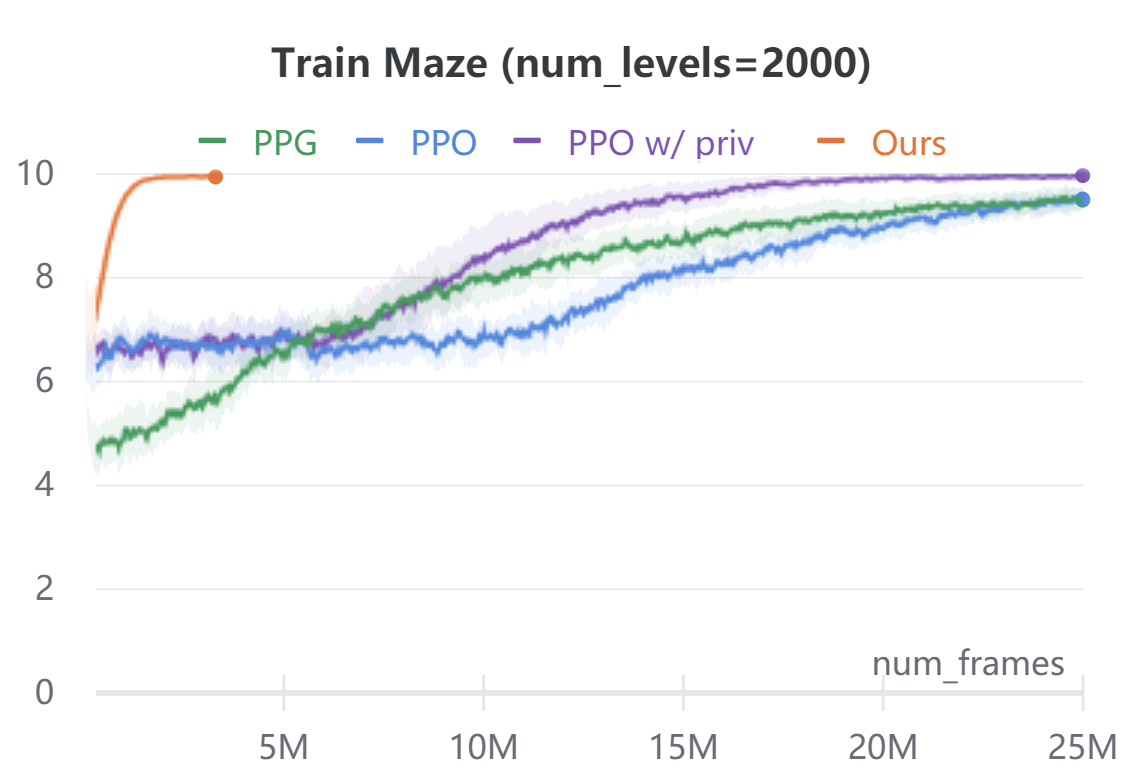}\caption{Maze 2000 training levels}\end{subfigure}%
\begin{subfigure}[b]{0.25\linewidth}\includegraphics[width=\linewidth]{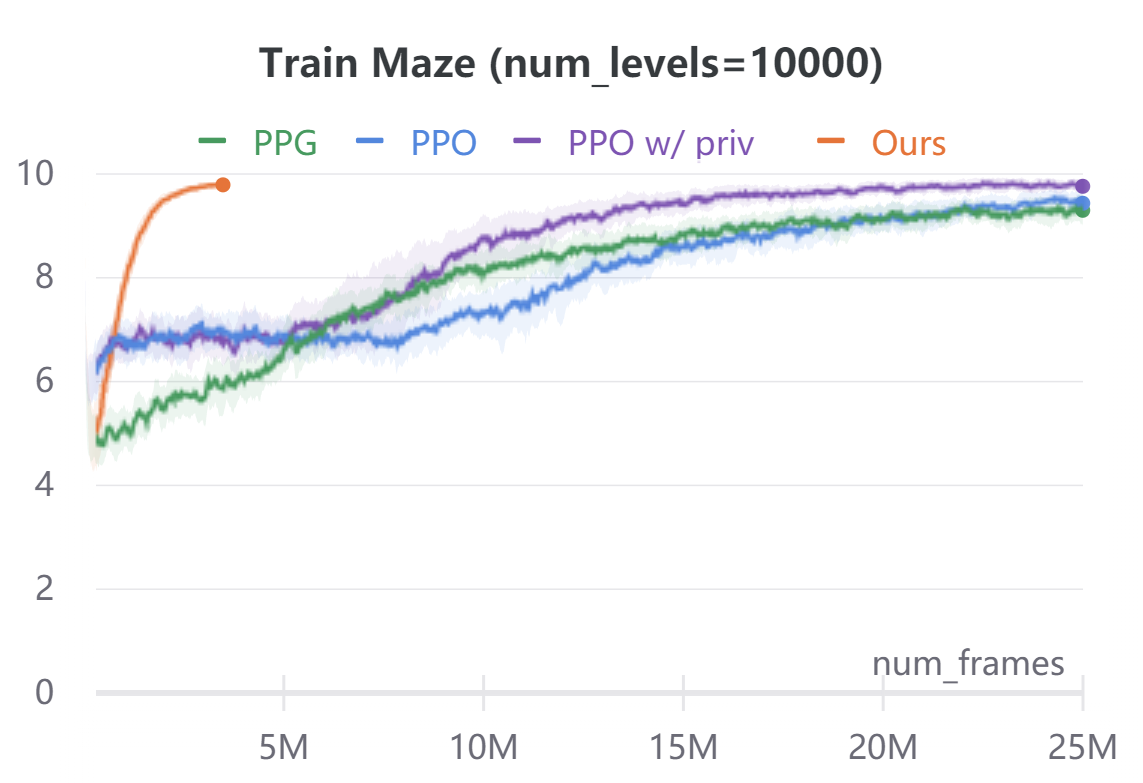}\caption{Maze 10000 training levels}\end{subfigure}%
\begin{subfigure}[b]{0.25\linewidth}\includegraphics[width=\linewidth]{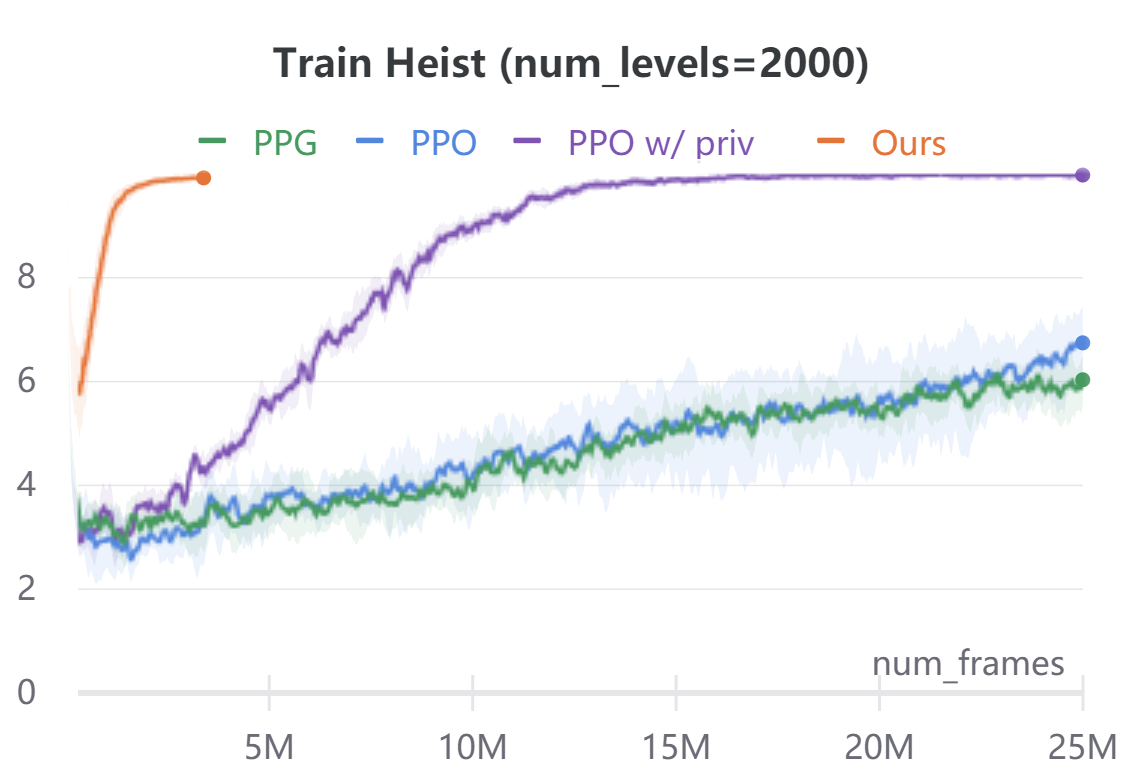}\caption{Heist 2000 training levels}\end{subfigure}%
\begin{subfigure}[b]{0.25\linewidth}\includegraphics[width=\linewidth]{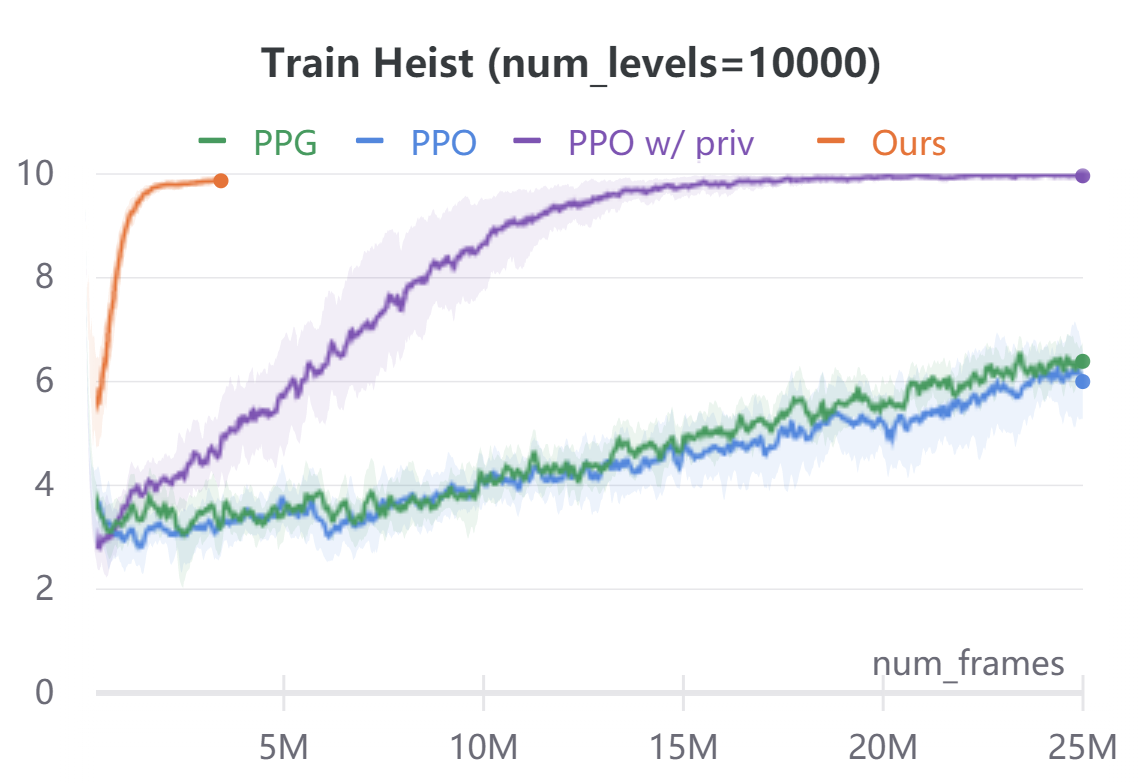}\caption{Heist 10000 training levels}\end{subfigure}%

\caption{Comparison of our method to state-of-the-art model-free reinforcement learning on the navigational tasks of the ProcGen benchmark.
All plots measure the average episode returns on the \textbf{training levels}.
Experimental setup follows \reffig{procgen_test}.}
\lblfig{procgen_train}
\end{figure*}

\section{Additional NoCrash Experiments}
\reftbl{nocrash_rc} additionally compares the route completion rates of the presented approach (\ourmethod) to prior state-of-the-art on the CARLA NoCrash benchmark.

\reftbl{noisy_visuomotor_option} compares a variant of our approach using noisy training trajectories (with Ornstein-Uhlenbeck noise~\cite{lillicrap2016continuous}).
This variant is most similar to data collected in LBC~\cite{chen2020learning}.
The experimental setup is equivalent to \reftbl{visuomotor_option}, just on noisy trajectories.
% Similar in \reftbl{visuomotor_option}, \textbf{CA} stands for camera augmentation, meaning the model trains on the additional augmented camera images, described in \refsec{dataset}. \textbf{SA} stands for ``speed augmentation''.

\begin{table}[t]
\centering
\begin{tabular}{l@{\ \ }c@{\ \ }c@{\ \ }c@{\ \ \ }c@{\ \ \ }c@{\ \ \ \ \ }c@{\ \ \ \ \ }}
\toprule
    Task && Town & Weather & IA & LBC & \textbf{\ourmethod} \\
\midrule
Empty &&\multirow{3}{*}{train} &\multirow{3}{*}{train} & $95.02$ & $97.15$ & $\bf{98.82}$ \\
Regular &&&& $94.72$ & $96.38$ & $\bf{100.00}$ \\
Dense &&&& $82.93$ & $91.35$ & $\bf{98.24}$ \\
\midrule
Empty &&\multirow{3}{*}{test} &\multirow{3}{*}{train} & $\it 88.87$ & $92.41$ & $\bf{98.91}$ \\
Regular &&&& $\it 84.09$ & $88.32$ & $\bf{94.95}$  \\
Dense &&&& $\it 63.63$ & $74.84$ & $\bf{88.89}$ \\
\midrule
Empty &&\multirow{3}{*}{train} &\multirow{3}{*}{test} & $-$ & $79.35$ & $\bf{94.25}$ \\
Regular &&&& $-$ & $79.20$ & $\bf{93.03}$ \\
Dense &&&& $-$ & $76.72$ & $\bf{95.73}$ \\
\midrule
Empty &&\multirow{3}{*}{test} &\multirow{3}{*}{test} & $-$ & $62.47$ & $\bf{84.72}$ \\
Regular &&&& $-$ & $63.55$ & $\bf{88.53}$ \\
Dense &&&& $-$ & $44.99$ & $\bf{80.75}$ \\
\bottomrule
\end{tabular}
\caption{Comparison of the \textbf{mean route completion rate} on NoCrash. The experimental setup follows \reftbl{nocrash}.}
\lbltbl{nocrash_rc}
\end{table}

\begin{table*}[t]
\centering
\begin{tabular}{c@{\ }c|c@{\ \ }c@{\ \ }c|c@{\ \ }c@{\ \ }c|c@{\ \ }c@{\ \ }c|c@{\ \ }c@{\ \ }c}
\toprule
    & & \multicolumn{6}{c|}{Train town} & \multicolumn{6}{c}{Test Town} \\
    & & \multicolumn{3}{c|}{Train Weather} & \multicolumn{3}{c|}{Test Weather} & \multicolumn{3}{c|}{Train Weather} & \multicolumn{3}{c}{Test Weather} \\
    CA & SA & Empty & Regular & Dense & Empty & Regular & Dense & Empty & Regular & Dense & Empty & Regular & Dense \\
\midrule
$\times$&$\times$&              $82$ & $88$ & $83$ & $70$ & $70$ & $64$ & $67$ & $72$ & $51$ & $44$ & $62$ & $40$\\
$\times$&$\checkmark$&          $95$ & $91$ & $\bf{97}$ & $90$ & $\bf{96}$ & $\bf{94}$ & $81$ & $80$ & $64$ & $54$ & $\bf{74}$ & $\bf{52}$ \\
$\checkmark$&$\times$&          $97$ & $96$ & $89$ & $84$ & $86$ & $74$ & $92$ & $87$ & $\bf{66}$ & $\bf{70}$ & $62$ & $36$ \\
$\checkmark$&$\checkmark$&$\bf{100}$ & $98$ & $94$ & $\bf{94}$ & $88$ & $82$ & $\bf{97}$ & $\bf{92}$ & $60$ & $64$ & $72$ & $46$ \\
\bottomrule
\end{tabular}
\caption{Comparison of success rate in the NoCrash benchmark trained on noisy trajectories, collected with injected Ornstein–Uhlenbeck~\cite{lillicrap2016continuous}.
Metric and evaluation protocol is comparable to \reftbl{visuomotor_option}, data-collection protocol follows LBC~\cite{chen2020learning}.}
\lbltbl{noisy_visuomotor_option}
\end{table*}

\section{Action-value Computation}
In CARLA, we use a planning horizon of $H=5$ to subsample the trajectories during action-value computation. At each frame $t$, we compute and discretize the rewards from $t$ to $t+H-1$ around the ego vehicle state at time $t$. We then compute the values and action-values for time $t$ using backward induction as described in section 3. In ProcGen, we use $H=30$.

\section{CARLA Controls}
In CARLA, to ensure a smooth control output from the discretized action space, we assume independence between steering and throttle, and use their softmax probabilities to compute smooth steering and throttle values. In particular, the sensorimotor policies predict logits $\log \pi_{s} \in \mathbb{R}^{N_s}, \log \pi_{t} \in \mathbb{R}^{N_t}, \log \pi_{b} \in \mathbb{R}$. During training we model $\log \pi(s,t,\mathbb{I}_{b}) = (1-\mathbb{I}_b)(\log \pi_s(s) + \log \pi_t(t)) + \mathbb{I}_b \log \pi_b$. During testing, we use 
\begin{align*}
    s &= \sum_{c}^{N_s}\pi_{s}(s_c)s_c \\
    t &= \sum_{c}^{N_t}\pi_{t}(t_c)s_c \\
    b &= \begin{cases}
        1, \pi_b >= t_b \\
        0, \pi_b <  t_b
    \end{cases}
\end{align*}
We use $t_b=0.5$ in all our experiments. In addition, we apply a bang-bang controller on throttle, i.e we explictly set the computed throttle to $0$ if the vehicle speed exceeds a predefined threshold.

\section{CARLA leaderboard}
Following~\citet{toromanoff2020end}, we use a 6 model ensemble to obtain a more stable control for our top leaderboard submission. 

\section{Training Hyperparameters}
\reftbl{hyperparams} provide a list of training hyperparameters for reference.
In our CARLA experiments we use the following image augmentations:
Gaussian Blur, 
Additive Gaussian Noise,
Pixel Dropout,
Multiply (scaling),
Linear Contrast,
Grayscale,
ElasticTransformation.

\begin{table}[t]
\centering
\begin{tabular}{lcc}
\toprule
Hyperparameter & CARLA & ProcGen \\
\midrule
Batch size & 128 & 128 \\
Learning rate - ego model & 1e-2 & 3e-4 \\
Learnign rate - distillation & 3e-4 & 3e-4 \\
Entropy loss scale ($\alpha$) & 1e-2 & 1e-2  \\
Segmentation loss scale & 5e-2 & $-$\\

\bottomrule
\end{tabular}
\caption{Additional hyperparameters.}
\lbltbl{hyperparams}
\end{table}

%\end{appendices}

% You must include your signed IEEE copyright release form when you submit
% your finished paper. We MUST have this form before your paper can be
% published in the proceedings.

\end{document}